\pdfoutput=1

\documentclass[11pt]{article}

\usepackage{naacl2021}
\usepackage{times}
\usepackage{latexsym}
\usepackage[T1]{fontenc}
\usepackage[utf8]{inputenc}
\usepackage{microtype}

\usepackage{booktabs}
\usepackage{graphicx}
\usepackage{multirow}
\usepackage{subcaption}
\usepackage{pifont}
\usepackage{enumitem}
\usepackage{pdflscape}
\usepackage{rotating}

\newcommand\blfootnote[1]{
  \begingroup
  \renewcommand\thefootnote{}\footnote{#1}
  \addtocounter{footnote}{-1}
  \endgroup
}

\title{A Survey on Recent Approaches for Natural Language Processing in Low-Resource Scenarios}

\author{Michael A. Hedderich*\textsuperscript{1}, Lukas Lange*\textsuperscript{1,2}, Heike Adel\textsuperscript{2}, \\ \textbf{Jannik Strötgen\textsuperscript{2} \& Dietrich Klakow\textsuperscript{1}}\\
\textsuperscript{1}Saarland University, Saarland Informatics Campus, Germany\\
\textsuperscript{2}Bosch Center for Artificial Intelligence, Germany\\
\texttt{\{mhedderich,dietrich.klakow\}@lsv.uni-saarland.de} \\
\texttt{\{lukas.lange,heike.adel,jannik.stroetgen\}@de.bosch.com}
}

\begin{document}
\maketitle
\begin{abstract}
Deep neural networks and huge language models are becoming omnipresent in natural language applications. As they are known for requiring large amounts of training data, there is a growing body of work to improve the performance in low-resource settings. Motivated by the recent fundamental changes towards neural models and the popular pre-train and fine-tune paradigm, we survey promising approaches for low-resource natural language processing. After a discussion about the different dimensions of data availability, we give a structured overview of methods that enable learning when training data is sparse. This includes mechanisms to create additional labeled data like data augmentation and distant supervision as well as transfer learning settings that reduce the need for target supervision. A goal of our survey is to explain how these methods differ in their requirements  as understanding them is essential for choosing a technique suited for a specific low-resource setting. Further key aspects of this work are to highlight open issues and to outline promising directions for future research.  
\end{abstract}

\section{Introduction}

\blfootnote{* equal contribution}
Most of today's research in natural language processing (NLP) is concerned with the processing of 10 to 20 high-resource languages with a special focus on English, and thus, ignores thousands of languages with billions of speakers~\cite{intro/bender2019rule}. 
The rise of data-hungry deep learning systems increased the performance of NLP for high resource-languages, but the shortage of large-scale data in less-resourced languages makes their processing a challenging problem.
Therefore, \newcite{intro/ruder2019problems} named NLP for low-resource scenarios one of the four biggest open problems in NLP nowadays.

The umbrella term low-resource covers a spectrum of scenarios with varying resource conditions. It includes work on threatened languages, such as Yongning Na, a Sino-Tibetan language with 40k speakers and only 3k written, unlabeled sentences \cite{emb/adams-etal-2017-cross}. Other languages are widely spoken but seldom addressed by NLP research. More than 310 languages exist with at least one million L1-speakers each \cite{intro/Ethnologue2019}. Similarly, Wikipedia exists for 300 languages.\footnote{\url{https://en.wikipedia.org/wiki/List_of_Wikipedias}} 
Supporting technological developments for low-resource languages can help to increase participation of the speakers' communities in a digital world. 
Note, however, that tackling low-resource settings is even crucial when dealing with popular NLP languages as low-resource settings do not only concern languages but also non-standard domains and tasks, for which -- even in English -- only little training data is available. Thus, the term ``language'' in this paper also includes domain-specific language. 

This importance of low-resource scenarios and the significant changes in NLP in the last years have led to active research on resource-lean settings and a wide variety of techniques have been proposed. They all share the motivation of overcoming the lack of labeled data by leveraging further sources. However, these works differ greatly on the sources they rely on, e.g., unlabeled data, manual heuristics or cross-lingual alignments. Understanding the requirements of these methods is essential for choosing a technique suited for a specific low-resource setting. Thus, one key goal of this survey is to highlight the underlying assumptions these techniques take regarding the low-resource setup.

\def\sectionautorefname{§}
\def\subsectionautorefname{§}

\begin{table*}
    \footnotesize
    \centering
    \begin{tabular}{p{4.2cm}|p{3.3cm}|p{3.75cm}|c|c}
    \toprule
    \multirow{2}{*}{\textbf{Method}} & \multirow{2}{*}{\textbf{Requirements}}  & \multirow{2}{*}{\textbf{Outcome}}  & \multicolumn{2}{c}{\textbf{For low-resource}} \\
    & & & \textbf{languages} & \textbf{domains}\\
    \midrule
    Data Augmentation (\autoref{sub:data-aug}) & labeled data, heuristics* & additional labeled data & \ding{51} & \ding{51} \\ \midrule
    Distant Supervision (\autoref{sub:distant}) & unlabeled data, heuristics* & additional labeled data & \ding{51} & \ding{51} \\ \midrule
    Cross-lingual  projections (\autoref{sub:projections}) & unlabeled data, high-resource labeled data, cross-lingual alignment & additional labeled data & \ding{51} & \ding{55} \\ \midrule
    Embeddings \& Pre-trained LMs (\autoref{sub:lm}) & unlabeled data & better language representation & \ding{51} & \ding{51} \\ \midrule
    LM domain adaptation (\autoref{sub:domain-lm}) & existing LM, \newline unlabeled domain data & domain-specific language representation & \ding{55} & \ding{51} \\ \midrule
    Multilingual LMs (\autoref{sub:multi-lm}) & multilingual unlabeled data & multilingual feature representation & \ding{51} & \ding{55} \\ \midrule
    Adversarial Discriminator (\autoref{sec:ml}) & additional datasets & independent representations & \ding{51} & \ding{51} \\ \midrule
    Meta-Learning (\autoref{sec:ml}) & multiple auxiliary tasks & better target task performance & \ding{51} & \ding{51}\\
    \bottomrule
    \end{tabular}
    \caption{Overview of low-resource methods surveyed in this paper. * Heuristics are typically gathered manually.}
    \label{tab:overview}
\end{table*}

In this work, we (1) give a broad and structured overview of current efforts on low-resource NLP, (2) analyse the different aspects of low-resource settings, (3) highlight the necessary resources and data assumptions as guidance for practitioners and (4) discuss open issues and promising future directions. Table \ref{tab:overview} gives an overview of the surveyed techniques along with their requirements a practitioner needs to take into consideration.

\section{Related Surveys}
\label{sec:surveys-etc}
Recent surveys cover low-resource machine translation \cite{survey/liu2019survey} and unsupervised domain adaptation \cite{survey/ramponi2020neural}. 
Thus, we do not investigate these topics further in this paper, but focus instead on general methods for low-resource, supervised natural language processing including data augmentation, distant supervision and transfer learning. This is also in contrast to the task-specific survey by \newcite{survey/magueresse2020low} who review highly influential work for several extraction tasks, but only provide little overview of recent approaches. In Table \ref{tab:surveys} in the appendix, we list past surveys that discuss a specific method or low-resource language family for those readers who seek a more specialized follow-up.

\section{Aspects of ``Low-Resource''}
\label{sec:defining}

\begin{figure}
    \centering
    \includegraphics[trim=55 80 190
60,clip,width=0.49\textwidth]{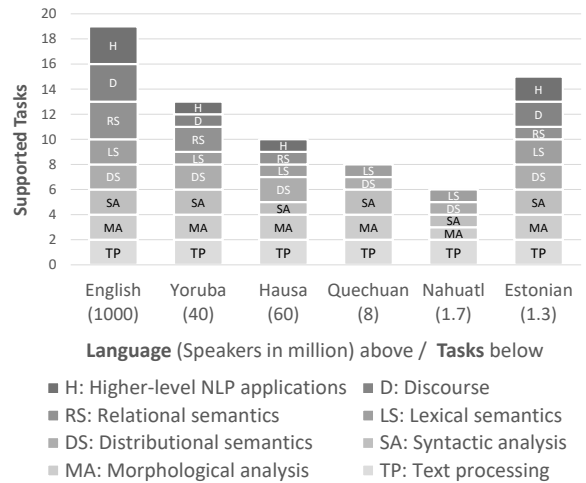}
    \caption{Supported NLP tasks in different languages. Note that the figure does not incorporate data quality or system performance. More details on the selection of tasks and languages are given in the appendix Section~\ref{app:tasks}.}
    \label{fig:applications}
\end{figure}

To visualize the variety of resource-lean scenarios, Figure~\ref{fig:applications} shows exemplarily which NLP tasks were addressed in six different languages from basic to higher-level tasks. While it is possible to build English NLP systems for many higher-level applications, low-resource languages lack the data foundation for this. Additionally, even if it is possible to create basic systems for tasks, such as tokenization and named entity recognition, for all tested low-resource languages, the training data is typical of lower quality compared to the English datasets, or very limited in size.
It also shows that the four American and African languages with between 1.5 and 60 million speakers have been addressed less than the Estonian language, with 1 million speakers. This indicates the unused potential to reach millions of speakers who currently have no access to higher-level NLP applications. \citet{survey/joshi2020linguisticdiversity} study further the availability of resources for languages around the world. 

\subsection{Dimensions of Resource Availability}

Many techniques presented in the literature depend on certain assumptions about the low-resource scenario. These have to be adequately defined to evaluate their applicability for a specific setting and to avoid confusion when comparing different approaches. 
We propose to categorize low-resource settings along the following three dimensions:

(i) The \textbf{availability of task-specific labels} in the target language (or target domain) is the most prominent dimension in the context of supervised learning. Labels are usually created through manual annotation, which can be both time- and cost-intensive. Not having access to adequate experts to perform the annotation can also be an issue for some languages and domains.

(ii) The  \textbf{availability of unlabeled language- or domain-specific text} is another factor, especially as most modern NLP approaches are based on some form of input embeddings trained on unlabeled texts. 

(iii) Most of the ideas surveyed in the next sections assume the \textbf{availability of auxiliary data} which can have many forms. Transfer learning might leverage task-specific labels in a different language or domain. Distant supervision utilizes external sources of information, such as knowledge bases or gazetteers. Some approaches require other NLP tools in the target language like machine translation to generate training data. It is essential to consider this as results from one low-resource scenario might not be transferable to another one if the assumptions on the auxiliary data are broken.

\subsection{How Low is Low-Resource?}
On the dimension of task-specific labels, different thresholds are used to define low-resource. 
For part-of-speech (POS) tagging, \newcite{distant/garrette2013pos2hours} limit the time of the annotators to 2 hours resulting in up to 1-2k tokens. \newcite{defining/kann2020POSPoorly} study languages that have less than 10k labeled tokens in the Universal Dependency project \cite{defining/Nivre2020UD} and  \newcite{survey/loubser2020viability} report that most available datasets for South African languages have 40-60k labeled tokens. 

The threshold is also task-dependent and more complex tasks might also increase the resource requirements. For text generation, \newcite{defining/yang2019responseGeneration} frame their work as low-resource with 350k labeled training instances. Similar to the task, the resource requirements can also depend on the language. \newcite{defining/plank2016MultilingualPOS} find that task performance varies between language families given the same amount of limited training data.

Given the lack of a hard threshold for low-resource settings, we see it as a spectrum of resource availability. We, therefore, also argue that more work should evaluate low-resource techniques across different levels of data availability for better comparison between approaches. 
For instance, \citet{defining/plank2016MultilingualPOS} and \citet{pre-train/melamud2019} show that 
for very small datasets non-neural methods outperform more modern approaches while the latter obtain better performance in resource-lean scenarios once a few hundred labeled instances are available.

\section{Generating Additional Labeled Data}
\label{sec:creating-additional-labeled-data}
Faced with the lack of task-specific labels, a variety of approaches have been developed to find alternative forms of labeled data as substitutes for gold-standard supervision. This is usually done through some form of expert insights in combination with automation. We group the ideas into two main categories: data augmentation which uses task-specific instances to create more of them (\autoref{sub:data-aug}) and distant supervision which labels unlabeled data (\autoref{sub:distant}) including cross-lingual projections (\autoref{sub:projections}). Additional sections cover learning with noisy labels (\autoref{sec:noisy}) and involving non-experts (\autoref{sec:nonexpert}).

\subsection{Data Augmentation}\label{sub:data-aug}
New instances can be obtained based on existing ones by modifying the features with transformations that do not change the label. In the computer vision community, this is a popular approach where, e.g., rotating an image is invariant to the classification of an image's content. For text, on the token level, this can be done by replacing words with equivalents, such as synonyms \cite{data-aug/wei-zou-2019-eda}, entities of the same type \cite{data-aug/raiman-miller-2017-globally,data-aug/dai2020analysis} or words that share the same morphology \cite{data-aug/gulordava2018Colorless,data-aug/vania2019parsing}. Such replacements can also be guided by a language model that takes context into consideration \cite{data-aug/fadaee-etal-2017-data,data-aug/kobayashi-2018-contextual}.

To go beyond the token level and add more diversity to the augmented sentences, data augmentation can also be performed on sentence parts. Operations that (depending on the task) do not change the label include manipulation of parts of the dependency tree \cite{data-aug/sahin2018morphing, data-aug/vania2019parsing,dehouck2020swapping}, simplification of sentences by removal of sentence parts \cite{data-aug/sahin2018morphing} and inversion of the subject-object relation \cite{data-aug/min-etal-2020-syntactic}. For whole sentences, paraphrasing through back-translation can be used. This is a popular approach in machine translation where target sentences are back-translated into source sentences \cite{data-aug/bojar2011backtranslation,data-aug/hoang2018iterativeBacktranslation}. An important aspect here is that errors in the source side/features do not seem to have a large negative effect on the generated target text the model needs to predict. It is therefore also used in other text generation tasks like abstract summarization \cite{data-aug/parida2019abstract} and table-to-text generation \cite{data-aug/ma2019table2text}. Back-translation has also been leveraged for text classification \cite{data-aug/xie2020backtranslation,data-aug/hegde2020paraphrase}. This setting assumes, however, the availability of a translation system. Instead, a language model can also be used for augmenting text classification datasets \cite{data-aug/kumar2020transformer-aug-gen, data-aug/anaby20notenough}. It is trained conditioned on a label, i.e., on the subset of the task-specific data with this label. It then generates additional sentences that fit this label. \newcite{data-aug/ding2020daga} extend this idea for token level tasks.

Adversarial methods are often used to find weaknesses in machine learning models \cite{adverserial/jin2020BERTrobust,adverserial/garg2020BAE}. They can, however, also be utilized to augment NLP datasets \cite{data-aug/yasunaga-etal-2018-robust,data-aug/morris2020textattack}. Instead of manually crafted transformation rules, these methods learn how to apply small perturbations to the input data that do not change the meaning of the text (according to a specific score). This approach is often applied on the level of vector representations. 
For instance, \newcite{data-aug/grundkiewicz2019Grammar} reverse the augmentation setting by applying transformations that flip the (binary) label. In their case, they introduce errors in correct sentences to obtain new training data for a grammar correction task. 

\textbf{Open Issues:} While data augmentation is ubiquitous in the computer vision community and while most of the above-presented approaches are task-independent, it has not found such widespread use in natural language processing. A reason might be that several of the approaches require an in-depth understanding of the language.  There is not yet a unified framework that allows applying data augmentation across tasks and languages. Recently, \citet{data-aug/longpre2020dataaug-vs-pretrain} hypothesised that data augmentation provides the same benefits as pre-training in transformer models. However, we argue that data augmentation might be better suited to leverage the insights of linguistic or domain experts in low-resource settings when unlabeled data or hardware resources are limited.

\subsection{Distant \& Weak Supervision}\label{sub:distant}

In contrast to data augmentation, distant or weak supervision uses unlabeled text and keeps it unmodified. The corresponding labels are obtained through a (semi-)automatic process from an external source of information. For named entity recognition (NER), a list of location names might be obtained from a dictionary and matches of tokens in the text with entities in the list are automatically labeled as locations. Distant supervision was introduced by \newcite{distant/mintz2009distant} for relation extraction (RE) with extensions on multi-instance \cite{distant/riedel10multi-instance} and multi-label learning \cite{distant/surdeanu-etal-2012-multi}. It is still a popular approach for information extraction tasks like NER and RE where the external information can be obtained from knowledge bases, gazetteers, dictionaries and other forms of structured knowledge sources \cite{distant/luo2017dynamicNoiseMatrix,distant/hedderich2018Noisy,distant/deng2019hop2,distant/alt2019finetuneOnDistant,distant/ye2019shift,distant/lange2019pharma,distant/nooralahzadeh2019reinforcementDenoising,distant/le2019entityLinking,distant/cao2019low,distant/lison-etal-2020-weak-supervision,distant/hedderich2021ANEA}. The automatic annotation ranges from simple string matching \cite{distant/yang2018distantPartialAnnotationReinforcement} to complex pipelines including classifiers and manual steps \cite{distant/norman2019distantMedical}. This distant supervision using information from external knowledge sources can be seen as a subset of the more general approach of labeling rules. These encompass also other ideas like reg-ex rules or simple programming functions \cite{distant/ratner2019snorkel,distant/zheng2019rules,survey/adelani2020distant,hedderich2020transfer,distant/lison-etal-2020-weak-supervision,ren2020denoising,karamanolakis2021leaving}. 

While distant supervision is popular for information extraction tasks like NER and RE, it is less prevalent in other areas of NLP. Nevertheless, distant supervision has also been successfully employed for other tasks by proposing new ways for automatic annotation. \newcite{distant/li-etal-2012-wiki} leverage a dictionary of POS tags for classifying unseen text with POS. For aspect classification, \citet{distant/karamanolakis-etal-2019-leveraging} create a simple bag-of-words classifier on a list of seed words and train a deep neural network on its weak supervision. \newcite{distant/wang2019sentiment} use context by transferring a document-level sentiment label to all its sentence-level instances. \newcite{distant/mekala2020meta} leverage meta-data for text classification and \newcite{huber2020discourse} build a discourse-structure dataset using guidance from sentiment annotations. For topic classification, heuristics can be used in combination with inputs from other classifiers like NER \cite{distant/bach19snorkeldrybell} or from entity lists \cite{hedderich2020transfer}. For some classification tasks, the labels can be rephrased with simple rules into sentences. A pre-trained language model then judges the label sentence that most likely follows the unlabeled input  \cite{distant/opitz19plausibility,distant/schick2020exploiting,distant/schick2020automatically}. An unlabeled review, for instance, might be continued with "It was great/bad" for obtaining binary sentiment labels.

\textbf{Open Issues:} 
The popularity of distant supervision for NER and RE might be due to these tasks being particularly suited. There, auxiliary data like entity lists is readily available and distant supervision often achieves reasonable results with simple surface form rules. It is an open question whether a task needs to have specific properties to be suitable for this approach. The existing work on other tasks and the popularity in other fields like image classification \cite{distant/xiao15clothing1m,distant/Li17webvision,distant/Lee18FoodNoisy,distant/mahajan18limits,distant/li20divide} suggests, however, that distant supervision could be leveraged for more NLP tasks in the future.

Distant supervision methods heavily rely on auxiliary data. In a low-resource setting, it might be difficult to obtain not only labeled data but also such auxiliary data.  \citet{defining/kann2020POSPoorly} find a large gap between the performance on high-resource and low-resource languages for POS tagging pointing to the lack of high-coverage and error-free dictionaries for the weak supervision in low-resource languages. This emphasizes the need for evaluating such methods in a realistic setting and avoiding to just simulate restricted access to labeled data in a high-resource language.

While distant supervision allows obtaining labeled data more quickly than manually annotating every instance of a dataset, it still requires human interaction to create automatic annotation techniques or to provide labeling rules. This time and effort could also be spent on annotating more gold label data, either naively or through an active learning scheme. Unfortunately, distant supervision papers rarely provide information on how long the creation took, making it difficult to compare these approaches. Taking the human expert into the focus connects this research direction with human-computer-interaction and human-in-the-loop setups \cite{distant/Klie18Inception,distant/qian-etal-2020-learning}. 

\subsection{Cross-Lingual Annotation Projections}\label{sub:projections}

For cross-lingual projections, a task-specific classifier is trained in a high-resource language. Using parallel corpora, the unlabeled low-resource data is then aligned to its equivalent in the high-resource language where labels can be obtained using the aforementioned classifier. These labels (on the high-resource text) can then be projected back to the text in the low-resource language based on the alignment between tokens in the parallel texts \cite{projection/yarowsky-etal-2001-inducing}. This approach can, therefore, be seen as a form of distant supervision specific for obtaining labeled data for low-resource languages. Cross-lingual projections have been applied in low-resource settings for tasks, such as POS tagging and parsing \cite{projections/tackstrom-etal-2013-token,projection/wisniewski-etal-2014-cross-lingual,projection/plank-agic-2018-distant,projection/eskander2020unsupervised}. 
Sources for parallel text can be the OPUS project \cite{projection/tiedemann-2012-parallel},
Bible corpora \cite{projection/mayer-cysouw-2014-creating,projections/Christodoulopoulos15Bible} or the recent JW300 corpus \cite{projection/agic-vulic-2019-jw300}.
Instead of using parallel corpora, existing high-resource labeled datasets can also be machine-translated into the low-resource language  \cite{projection/khalil-etal-2019-crossIntent,projection/zhang-etal-2019-cross,projection/fei-etal-2020-cross,projection/amjad2020mt}. Cross-lingual projections have even been used with English as a target language for detecting linguistic phenomena like modal sense and telicity that are easier to identify in a different language \cite{projection/zhou-etal-2015-semantically,projection/marasovic-etal-2016-modal,projection/friedrich-gateva-2017telicity}.

\textbf{Open issues:} Cross-lingual projections set high requirements on the auxiliary data needing both labels in a high-resource language and means to project them into a low-resource language. Especially the latter can be an issue as machine translation by itself might be problematic for a specific low-resource language. A limitation of the parallel corpora is their domains like political proceedings or religious texts. \newcite{projection/mayhew-etal-2017-cheap}, \newcite{projection/fang-cohn-2017-model} and \citet{cross-ling/karamanolakis2020seedwords} propose systems with fewer requirements based on word translations, bilingual dictionaries and task-specific seed words, respectively. 

\subsection{Learning with Noisy Labels}
\label{sec:noisy}
The above-presented methods allow obtaining labeled data quicker and cheaper than manual annotations. These labels tend, however, to contain more errors. Even though more training data is available, training directly on this noisily-labeled data can actually hurt the performance. Therefore, many recent approaches for distant supervision use a noise handling method to diminish the negative effects of distant supervision. We categorize these into two ideas: noise filtering and noise modeling.

Noise filtering methods remove instances from the training data that have a high probability of being incorrectly labeled. This often includes training a classifier to make the filtering decision. The filtering can remove the instances completely from the training data, e.g., through a probability threshold \cite{distant/jia2019Arnor}, a binary classifier \cite{distant/adel15coreference,distant/onoe2019filteringRelabeling,distant/huang2019curriculum}, or the use of a reinforcement-based agent \cite{distant/yang2018distantPartialAnnotationReinforcement,distant/nooralahzadeh2019reinforcementDenoising}. Alternatively, a soft filtering might be applied that re-weights instances according to their probability of being correctly labeled \cite{distant/le2019entityLinking} or an attention measure \cite{distant/hu2019jointLabel}.

The noise in the labels can also be modeled. A common model is a confusion matrix estimating the relationship between clean and noisy labels \cite{distant/fang-cohn-2016-learning,distant/luo2017dynamicNoiseMatrix,distant/hedderich2018Noisy,distant/paul-etal-2019-handling,distant/lange2019pharma,distant/lange2019feature,distant/chen-etal-2019-uncover,distant/wang2019sentiment,distant/hedderich21noisemodel}. The classifier is no longer trained directly on the noisily-labeled data. Instead, a noise model is appended which shifts the noisy to the (unseen) clean label distribution. This can be interpreted as the original classifier being trained on a ``cleaned'' version of the noisy labels. In \newcite{distant/ye2019shift}, the prediction is shifted from the noisy to the clean distribution during testing. In \newcite{distant/chen-etal-2020-relabel-agents}, a group of reinforcement agents relabels noisy instances. \newcite{distant/rehbein-ruppenhofer-2017-detecting}, \newcite{distant/lison-etal-2020-weak-supervision} and \newcite{ren2020denoising} leverage several sources of distant supervision and learn how to combine them.

In NER, the noise in distantly supervised labels tends to be false negatives,
i.e., mentions of entities that have been missed by the automatic method. Partial annotation learning \cite{distant/yang2018distantPartialAnnotationReinforcement,distant/nooralahzadeh2019reinforcementDenoising,distant/cao2019low} takes this into account explicitly. Related approaches learn latent variables \cite{distant/jie-etal-2019-betterIncomplete}, use constrained binary learning \cite{distant/mayhew2019partiallyNonSpeaker} or construct a loss assuming  that only unlabeled positive instances exist \cite{distant/peng-etal-2019-positive-unlabeled}. 

\subsection{Non-Expert Support}
\label{sec:nonexpert}
As an alternative to an automatic annotation process, annotations might also be provided by non-experts. Similar to distant supervision, this results in a trade-off between label quality and availability. For instance, \newcite{distant/garrette2013pos2hours} obtain labeled data from non-native-speakers and without a quality control on the manual annotations. This can be taken even further by employing annotators who do not speak the low-resource language \cite{distant/mayhew2018talen,distant/mayhew2019partiallyNonSpeaker,distant/tsygankova2020nonspeaker}. 

\citet{nekoto2020participatory} take the opposite direction, integrating speakers of low-resource languages without formal training into the model development process in an approach of participatory research. This is part of recent work on how to strengthen low-resource language communities and grassroot approaches \cite{distant/alnajjar2020verdd,adelani2021MasakhaNER}. 

\section{Transfer Learning} 
\label{sec:transferlearning}
While distant supervision and data augmentation generate and extend task-specific training data, transfer learning reduces the need for labeled target data by transferring learned representations and models.
A strong focus in recent works on transfer learning in NLP lies in the use of pre-trained language representations that are trained on unlabeled data like BERT~\cite{pre-train/devlin2019bert}. Thus, this section starts with an overview of these methods (\autoref{sub:lm}) and then discusses how they can be utilized in low-resource scenarios, in particular, regarding the usage in domain-specific (\autoref{sub:domain-lm}) or multilingual low-resource settings (\autoref{sub:multi-lm}).

\subsection{Pre-Trained Language Representations}\label{sub:lm}
Feature vectors are the core input component of many neural network-based models for NLP tasks. They are numerical representations of words or sentences, as neural architectures do not allow the processing of strings and characters as such. \newcite{pre-train/collobert2011natural} showed that training these models for the task of language-modeling on a large-scale corpus results in high-quality word representations, which can be reused for other downstream tasks as well.
Subword-based embeddings such as fastText n-gram embeddings~\cite{emb/bojanowski2017enriching} and byte-pair-encoding embeddings \cite{emb/heinzerling2018bpemb} addressed out-of-vocabulary issues by splitting words into multiple subwords, which in combination represent the original word. 
\newcite{emb/zhu2019subword} showed that these embeddings leveraging subword information are beneficial for low-resource sequence labeling tasks, such as named entity recognition and typing, and outperform word-level embeddings.  \newcite{emb/jungmaier2020dirichlet} added smoothing to word2vec models to correct its bias towards rare words and achieved improvements in particular for low-resource settings.
In addition, pre-trained embeddings were published for more than 270 languages for both embedding methods.
This enabled the processing of texts in many languages, including multiple low-resource languages found in Wikipedia.
More recently, a trend emerged of pre-training large embedding models using a language model objective to create context-aware word representations by predicting the next word or sentence. 
This includes pre-trained transformer models~\cite{pre-train/vaswani2017attention}, such as BERT~\cite{pre-train/devlin2019bert} or RoBERTa~\cite{pre-train/liu2019roberta}. 
These methods are particularly helpful for low-resource languages for which large amounts of unlabeled data are available, but task-specific labeled data is scarce~\cite{pre-train/cruz2019evaluating}.

\textbf{Open Issues:} While pre-trained language models achieve significant performance increases compared to standard word embeddings, it is still questionable if these methods are suited for real-world low-resource scenarios. 
For example, all of these models require large hardware requirements, in particular, considering that the transformer model size keeps increasing to boost performance~\cite{pre-train/raffel2019exploring}.
Therefore, these large-scale methods might not be suited for low-resource scenarios where hardware is also low-resource.\\
\newcite{pre-train/van2020optimal} showed that low- to medium-depth transformer sizes perform better than larger models for low-resource languages and \newcite{bert/schick2020s} managed to train models with three orders of magnitude fewer parameters that perform on-par with large-scale models like GPT-3 on few-shot task by reformulating the training task and using ensembling.
\newcite{pre-train/melamud2019} showed that simple bag-of-words approaches are better when there are only a few dozen training instances or less for text classification, while more complex transformer models require more training data.
\citet{bert/bhattacharjee-etal-2020-bert} found that cross-view training \cite{cvt/clark-etal-2018-semi} leverages large amounts of unlabeled data better for task-specific applications in contrast to the general representations learned by BERT.
Moreover, data quality for low-resource, even for unlabeled data, might not be comparable to data from high-resource languages. \newcite{defining/alabi-etal-2020-massive} found that word embeddings trained on larger amounts of unlabeled data from low-resource languages are not competitive to embeddings trained on smaller, but curated data sources.

\subsection{Domain-Specific Pre-Training}\label{sub:domain-lm}
The language of a specialized domain can differ tremendously from what is considered the standard language, thus, many text domains are often less-resourced as well. For example, scientific articles can contain formulas and technical terms, which are not observed in news articles. 
However, the majority of recent language models are pre-trained on general-domain data, such as texts from the news or web-domain, which can lead to a so-called ``domain-gap'' when applied to a different domain.

One solution to overcome this gap is the adaptation to the target domain by finetuning the language model.
\newcite{domain/gugurangan2020dontstop} showed that continuing the training of a model with additional domain-adaptive and task-adaptive pre-training with unlabeled data leads to performance gains for both high- and low-resource settings for numerous English domains and tasks.
This is also displayed in the number of domain-adapted language models \cite[(i.a.)]{domain/alsentzer2019publicly,domain/huang2019clinicalbert,domain/adhikari2019docbert,domain/lee2020patent,domain/jain2020nukebert}, 
most notably BioBERT~\cite{domain/lee2020biobert} that was pre-trained on biomedical PubMED articles and SciBERT~\cite{domain/beltagy2019scibert} for scientific texts. 
For example, \citet{domain/friedrich2020sofc} showed that a general-domain BERT model performs well in the materials science domain, but the domain-adapted SciBERT performs best. 
\newcite{domain/xu2020dombert} used in- and out-of-domain data to pre-train a domain-specific model and adapt it to low-resource domains.
\newcite{domain/aharoni2020unsupervised} found domain-specific clusters in pre-trained language models and showed how these could be exploited for data selection in domain-sensitive training. 

Powerful representations can be achieved by combining high-resource embeddings from the general domain with low-resource embeddings from the target domain \cite{domain/akbik2018contextual, domain/lange2019nlnde}. 
\newcite{emb/kiela-etal-2018-dynamic} showed that embeddings from different domains can be combined using attention-based meta-embeddings, which create a weighted sum of all embeddings. \newcite{emb/lange2020adversarial} further improved on this by aligning embeddings trained on diverse domains using an adversarial discriminator that distinguishes between the embedding spaces to generate domain-invariant representations.

\subsection{Multilingual Language Models}\label{sub:multi-lm}
Analogously to low-resource domains, low-resource languages can also benefit from labeled resources available in other high-resource languages.
This usually requires the training of multilingual language representations by combining monolingual representations \cite{emb/lange-etal-2020-choice} or training a single model for many languages, such as multilingual BERT~\cite{pre-train/devlin2019bert} or XLM-RoBERTa~\cite{pre-train/conneau2020unsupervised} .
These models are trained using unlabeled, monolingual corpora from different languages and can be used in cross- and multilingual settings, due to many languages seen during pre-training. 

In cross-lingual zero-shot learning, no task-specific labeled data is available in the low-resource target language. Instead, labeled data from a  high-resource language is leveraged. 
A multilingual model can be trained on the target task in a high-resource language and afterwards, applied to the unseen target languages, such as for named entity recognition \cite{cross-ling/lin-etal-2019-choosing,cross-ling/hvingelby-etal-2020-dane}, reading comprehension \cite{cross-ling/hsu-etal-2019-zero},  temporal expression extraction \cite{cross-ling/lange2020adversarial}, or POS tagging and dependency parsing \cite{mueller20multitransfer}.
\citet{eval/hu2020xtreme} showed, however, that there is still a large gap between low and high-resource setting.
\citet{lauscher2020zero} and \citet{hedderich2020transfer} proposed adding a minimal amount of target-task and -language data (in the range of 10 to 100 labeled sentences) which resulted in a  significant boost in performance for classification in low-resource languages.

The transfer between two languages can be improved by creating a common multilingual embedding space of multiple languages. 
This is useful for standard word embeddings \cite{cross-ling/ruder2019survey} as well as pre-trained language models. For example, by aligning the languages inside a single multilingual model, i.a., in cross-lingual~\cite{cross-ling/schuster2019cross,cross-ling/liu2019investigating} or multilingual settings \cite{multi-ling/cao2020align}. 

This alignment is typically done by computing a mapping between two different embedding spaces, such that the words in both embeddings share similar feature vectors after the mapping \cite{multi-ling/mikolov2013exploiting,multi-ling/joulin2018loss}. 
This allows to use different embeddings inside the same model and helps when two languages do not share the same space inside a single model \cite{multi-ling/cao2020align}.
For example, \newcite{cross-ling/zhang2019improving} used bilingual representations by creating cross-lingual word embeddings using a small set of parallel sentences between the high-resource language English and three low-resource African languages, Swahili, Tagalog, and Somali, to improve document retrieval performance for the African languages.

\begin{figure}
    \centering
    \includegraphics[trim=60 330 60
60,clip,width=0.49\textwidth]{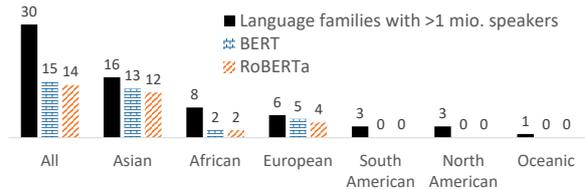}
    \caption{Language families with more than 1 million speakers covered by multilingual transformer models. }
    \label{fig:language_families}
\end{figure}

\textbf{Open Issues:} While these multilingual models are a tremendous step towards enabling NLP in many languages, possible claims that these are universal language models do not hold. 
For example, mBERT covers 104 and XLM-R 100 languages,
which is a third of all languages in Wikipedia as outlined earlier.  
Further, \newcite{multi-ling/wu-dredze-2020-languages} showed that, in particular, low-resource languages are not well-represented in mBERT.
Figure~\ref{fig:language_families} shows which language families with at least 1 million speakers are covered by mBERT and XLM-RoBERTa\footnote{A language family is covered if at least one associated language is covered. Language families can belong to multiple regions, e.g., Indo-European belongs to Europe and Asia.}. 
In particular, African and American languages are not well-represented within the transformer models, even though millions of people speak these languages.
This can be problematic, as languages from more distant language families are less suited for transfer learning, as \newcite{lauscher2020zero} showed.

\section{Ideas From Low-Resource Machine Learning in Non-NLP Communities}\label{sec:ml}
Training on a limited amount of data is not unique to natural language processing. Other areas, like general machine learning and computer vision, can be a useful source for insights and new ideas. We already presented data augmentation and pre-training. Another example is Meta-Learning \cite{transfer/Finn17Meta}, which is based on multi-task learning. Given a set of auxiliary high-resource tasks and a low-resource target task, meta-learning trains a model to decide how to use the auxiliary tasks in the most beneficial way for the target task. For NLP, this approach has been evaluated on tasks such as sentiment analysis \cite{transfer/yu2018metrics}, user intent classification \cite{transfer/yu2018metrics,transfer/chen2020domain}, natural language understanding \cite{transfer/dou2019meta}, text classification \cite{bansal-etal-2020-self} and dialogue generation \cite{transfer/huang2020meta}. Instead of having a set of tasks, \newcite{transfer/rahimi-etal-2019-massively} built an ensemble of language-specific NER models which are then weighted depending on the zero- or few-shot target language.

Differences in the features between the pre-training and the target domain can be an issue in transfer learning, especially in neural approaches where it can be difficult to control which information the model takes into account. Adversarial discriminators \cite{transfer/goodfellow2014gan} can prevent the model from learning a feature-representation that is specific to a data source. \newcite{transfer/gui-etal-2017-part-adversarial}, \newcite{transfer/liu2017adversarial}, \newcite{transfer/kasai2019adversarial},  \newcite{domain/griesshaber2020low} and
 \newcite{transfer/zhou2019dualAdversarial} learned domain-independent representations using adversarial training. 
 \newcite{transfer/kim-etal-2017-cross-adversarial}, \newcite{domain/chen2018adversarial} and \newcite{cross-ling/lange2020adversarial} worked with language-independent representations for cross-lingual transfer. These examples show the beneficial exchange of ideas between NLP and the machine learning community.

\section{Discussion and Conclusion}

In this survey, we gave a structured overview of recent work in the field of low-resource natural language processing. Beyond the method-specific open issues presented in the previous sections, we see the comparison between approaches as an important point of future work. Guidelines are necessary to support practitioners in choosing the right tool for their task. In this work, we highlighted that it is essential to analyze resource-lean scenarios across the different dimensions of data-availability. This can reveal which techniques are expected to be applicable in a specific low-resource setting. More theoretic and experimental work is necessary to understand how approaches compare to each other and on which factors their effectiveness depends. \citet{data-aug/longpre2020dataaug-vs-pretrain}, for instance, hypothesized that data augmentation and pre-trained language models yield similar kind of benefits. Often, however, new techniques are just compared to similar methods and not across the range of low-resource approaches. While a fair comparison is non-trivial given the different requirements on auxiliary data, we see this endeavour as essential to improve the field of low-resource learning in the future. This could also help to understand where the different approaches complement each other and how they can be combined effectively.

\section*{Acknowledgments}
The authors would like to thank Annemarie Friedrich for her valuable feedback and the anonymous reviewers for their helpful comments.
This work has been partially funded by the Deutsche Forschungsgemeinschaft (DFG, German Research Foundation) – Project-ID 232722074 – SFB 1102 and the EU Horizon 2020 project ROXANNE under grant number 833635.
\bibliography{survey}

\begin{thebibliography}{220}
\expandafter\ifx\csname natexlab\endcsname\relax\def\natexlab#1{#1}\fi

\bibitem[{Adams et~al.(2017)Adams, Makarucha, Neubig, Bird, and
  Cohn}]{emb/adams-etal-2017-cross}
Oliver Adams, Adam Makarucha, Graham Neubig, Steven Bird, and Trevor Cohn.
  2017.
\newblock \href {https://www.aclweb.org/anthology/E17-1088} {Cross-lingual word
  embeddings for low-resource language modeling}.
\newblock In \emph{Proceedings of the 15th Conference of the {E}uropean Chapter
  of the Association for Computational Linguistics: Volume 1, Long Papers},
  pages 937--947, Valencia, Spain. Association for Computational Linguistics.

\bibitem[{Adel and Sch{\"{u}}tze(2015)}]{distant/adel15coreference}
Heike Adel and Hinrich Sch{\"{u}}tze. 2015.
\newblock \href
  {https://tac.nist.gov/publications/2015/participant.papers/TAC2015.CIS.proceedings.pdf}
  {{CIS} at {TAC} cold start 2015: Neural networks and coreference resolution
  for slot filling}.
\newblock In \emph{Proceedings of TAC KBP Workshop}.

\bibitem[{Adelani et~al.(2021)Adelani, Abbott, Neubig, D'souza, Kreutzer,
  Lignos, Palen-Michel, Buzaaba, Rijhwani, Ruder, Mayhew, Azime, Muhammad,
  Emezue, Nakatumba-Nabende, Ogayo, Aremu, Gitau, Mbaye, Alabi, Yimam, Gwadabe,
  Ezeani, Niyongabo, Mukiibi, Otiende, Orife, David, Ngom, Adewumi, Rayson,
  Adeyemi, Muriuki, Anebi, Chukwuneke, Odu, Wairagala, Oyerinde, Siro, Bateesa,
  Oloyede, Wambui, Akinode, Nabagereka, Katusiime, Awokoya, MBOUP,
  Gebreyohannes, Tilaye, Nwaike, Wolde, Faye, Sibanda, Ahia, Dossou, Ogueji,
  DIOP, Diallo, Akinfaderin, Marengereke, and Osei}]{adelani2021MasakhaNER}
David~Ifeoluwa Adelani, Jade Abbott, Graham Neubig, Daniel D'souza, Julia
  Kreutzer, Constantine Lignos, Chester Palen-Michel, Happy Buzaaba, Shruti
  Rijhwani, Sebastian Ruder, Stephen Mayhew, Israel~Abebe Azime, Shamsuddeen
  Muhammad, Chris~Chinenye Emezue, Joyce Nakatumba-Nabende, Perez Ogayo,
  Anuoluwapo Aremu, Catherine Gitau, Derguene Mbaye, Jesujoba Alabi, Seid~Muhie
  Yimam, Tajuddeen Gwadabe, Ignatius Ezeani, Rubungo~Andre Niyongabo, Jonathan
  Mukiibi, Verrah Otiende, Iroro Orife, Davis David, Samba Ngom, Tosin Adewumi,
  Paul Rayson, Mofetoluwa Adeyemi, Gerald Muriuki, Emmanuel Anebi, Chiamaka
  Chukwuneke, Nkiruka Odu, Eric~Peter Wairagala, Samuel Oyerinde, Clemencia
  Siro, Tobius~Saul Bateesa, Temilola Oloyede, Yvonne Wambui, Victor Akinode,
  Deborah Nabagereka, Maurice Katusiime, Ayodele Awokoya, Mouhamadane MBOUP,
  Dibora Gebreyohannes, Henok Tilaye, Kelechi Nwaike, Degaga Wolde, Abdoulaye
  Faye, Blessing Sibanda, Orevaoghene Ahia, Bonaventure F.~P. Dossou, Kelechi
  Ogueji, Thierno~Ibrahima DIOP, Abdoulaye Diallo, Adewale Akinfaderin, Tendai
  Marengereke, and Salomey Osei. 2021.
\newblock \href {https://arxiv.org/pdf/2103.11811.pdf} {Masakhaner: Named
  entity recognition for african languages}.
\newblock \emph{arXiv preprint arXiv:2103.11811}.

\bibitem[{Adelani et~al.(2020)Adelani, Hedderich, Zhu, Berg, and
  Klakow}]{survey/adelani2020distant}
David~Ifeoluwa Adelani, Michael~A Hedderich, Dawei Zhu, Esther van~den Berg,
  and Dietrich Klakow. 2020.
\newblock \href {https://arxiv.org/pdf/2003.08370.pdf} {Distant supervision and
  noisy label learning for low resource named entity recognition: A study on
  hausa and yoruba}.
\newblock \emph{Workshop on Practical Machine Learning for Developing Countries
  at ICLR'20}.

\bibitem[{Adhikari et~al.(2019)Adhikari, Ram, Tang, and
  Lin}]{domain/adhikari2019docbert}
Ashutosh Adhikari, Achyudh Ram, Raphael Tang, and Jimmy Lin. 2019.
\newblock \href {https://arxiv.org/pdf/1904.08398.pdf} {Docbert: Bert for
  document classification}.
\newblock \emph{arXiv preprint arXiv:1904.08398}.

\bibitem[{Aggarwal et~al.(2014)Aggarwal, Kong, Gu, Han, and
  Philip}]{active/aggarwal2014survey}
Charu~C Aggarwal, Xiangnan Kong, Quanquan Gu, Jiawei Han, and S~Yu Philip.
  2014.
\newblock \href {http://charuaggarwal.net/active-survey.pdf} {Active learning:
  A survey}.
\newblock In \emph{Data Classification: Algorithms and Applications}, pages
  571--605. CRC Press.

\bibitem[{Agi{\'c} and Vuli{\'c}(2019)}]{projection/agic-vulic-2019-jw300}
{\v{Z}}eljko Agi{\'c} and Ivan Vuli{\'c}. 2019.
\newblock \href {https://doi.org/10.18653/v1/P19-1310} {{JW}300: A
  wide-coverage parallel corpus for low-resource languages}.
\newblock In \emph{Proceedings of the 57th Annual Meeting of the Association
  for Computational Linguistics}, pages 3204--3210, Florence, Italy.
  Association for Computational Linguistics.

\bibitem[{Aharoni and Goldberg(2020)}]{domain/aharoni2020unsupervised}
Roee Aharoni and Yoav Goldberg. 2020.
\newblock \href {https://www.aclweb.org/anthology/2020.acl-main.692}
  {Unsupervised domain clusters in pretrained language models}.
\newblock In \emph{Proceedings of the 58th Annual Meeting of the Association
  for Computational Linguistics}, pages 7747--7763, Online. Association for
  Computational Linguistics.

\bibitem[{Akbik et~al.(2018)Akbik, Blythe, and
  Vollgraf}]{domain/akbik2018contextual}
Alan Akbik, Duncan Blythe, and Roland Vollgraf. 2018.
\newblock \href {https://www.aclweb.org/anthology/C18-1139} {Contextual string
  embeddings for sequence labeling}.
\newblock In \emph{Proceedings of the 27th International Conference on
  Computational Linguistics}, pages 1638--1649, Santa Fe, New Mexico, USA.
  Association for Computational Linguistics.

\bibitem[{Al-Ayyoub et~al.(2018)Al-Ayyoub, Nuseir, Alsmearat, Jararweh, and
  Gupta}]{survey/al2018deep}
Mahmoud Al-Ayyoub, Aya Nuseir, Kholoud Alsmearat, Yaser Jararweh, and Brij
  Gupta. 2018.
\newblock \href
  {https://www.sciencedirect.com/science/article/pii/S1877750317303757} {Deep
  learning for arabic nlp: A survey}.
\newblock \emph{Journal of computational science}, 26:522--531.

\bibitem[{Alabi et~al.(2020)Alabi, Amponsah-Kaakyire, Adelani, and
  Espa{\~n}a-Bonet}]{defining/alabi-etal-2020-massive}
Jesujoba Alabi, Kwabena Amponsah-Kaakyire, David Adelani, and Cristina
  Espa{\~n}a-Bonet. 2020.
\newblock \href {https://www.aclweb.org/anthology/2020.lrec-1.335} {Massive vs.
  curated embeddings for low-resourced languages: the case of yor{\`u}b{\'a}
  and twi}.
\newblock In \emph{Proceedings of The 12th Language Resources and Evaluation
  Conference}, pages 2754--2762, Marseille, France. European Language Resources
  Association.

\bibitem[{Algan and Ulusoy(2021)}]{survey/algan2019image}
G{\"o}rkem Algan and Ilkay Ulusoy. 2021.
\newblock \href
  {https://www.sciencedirect.com/science/article/pii/S0950705121000344} {Image
  classification with deep learning in the presence of noisy labels: A survey}.
\newblock \emph{Knowledge-Based Systems}, 215:106771.

\bibitem[{Alnajjar et~al.(2020)Alnajjar, H{\"a}m{\"a}l{\"a}inen, Rueter, and
  Partanen}]{distant/alnajjar2020verdd}
Khalid Alnajjar, Mika H{\"a}m{\"a}l{\"a}inen, Jack Rueter, and Niko Partanen.
  2020.
\newblock \href {https://doi.org/10.18653/v1/2020.coling-demos.1} {Ve{'}rdd.
  narrowing the gap between paper dictionaries, low-resource {NLP} and
  community involvement}.
\newblock In \emph{Proceedings of the 28th International Conference on
  Computational Linguistics: System Demonstrations}, pages 1--6, Barcelona,
  Spain (Online). International Committee on Computational Linguistics (ICCL).

\bibitem[{Alsentzer et~al.(2019)Alsentzer, Murphy, Boag, Weng, Jindi, Naumann,
  and McDermott}]{domain/alsentzer2019publicly}
Emily Alsentzer, John Murphy, William Boag, Wei-Hung Weng, Di~Jindi, Tristan
  Naumann, and Matthew McDermott. 2019.
\newblock \href {https://doi.org/10.18653/v1/W19-1909} {Publicly available
  clinical {BERT} embeddings}.
\newblock In \emph{Proceedings of the 2nd Clinical Natural Language Processing
  Workshop}, pages 72--78, Minneapolis, Minnesota, USA. Association for
  Computational Linguistics.

\bibitem[{Alt et~al.(2019)Alt, H{\"u}bner, and
  Hennig}]{distant/alt2019finetuneOnDistant}
Christoph Alt, Marc H{\"u}bner, and Leonhard Hennig. 2019.
\newblock \href {https://doi.org/10.18653/v1/P19-1134} {Fine-tuning pre-trained
  transformer language models to distantly supervised relation extraction}.
\newblock In \emph{Proceedings of the 57th Annual Meeting of the Association
  for Computational Linguistics}, pages 1388--1398, Florence, Italy.
  Association for Computational Linguistics.

\bibitem[{Amjad et~al.(2020)Amjad, Sidorov, and Zhila}]{projection/amjad2020mt}
Maaz Amjad, Grigori Sidorov, and Alisa Zhila. 2020.
\newblock \href {https://www.aclweb.org/anthology/2020.lrec-1.309} {Data
  augmentation using machine translation for fake news detection in the {U}rdu
  language}.
\newblock In \emph{Proceedings of the 12th Language Resources and Evaluation
  Conference}, pages 2537--2542, Marseille, France. European Language Resources
  Association.

\bibitem[{Anaby{-}Tavor et~al.(2020)Anaby{-}Tavor, Carmeli, Goldbraich, Kantor,
  Kour, Shlomov, Tepper, and Zwerdling}]{data-aug/anaby20notenough}
Ateret Anaby{-}Tavor, Boaz Carmeli, Esther Goldbraich, Amir Kantor, George
  Kour, Segev Shlomov, Naama Tepper, and Naama Zwerdling. 2020.
\newblock \href {https://aaai.org/ojs/index.php/AAAI/article/view/6233} {Do not
  have enough data? deep learning to the rescue!}
\newblock In \emph{The Thirty-Fourth {AAAI} Conference on Artificial
  Intelligence, {AAAI} 2020, The Thirty-Second Innovative Applications of
  Artificial Intelligence Conference, {IAAI} 2020, The Tenth {AAAI} Symposium
  on Educational Advances in Artificial Intelligence, {EAAI} 2020, New York,
  NY, USA, February 7-12, 2020}, pages 7383--7390. {AAAI} Press.

\bibitem[{Bach et~al.(2019)Bach, Rodriguez, Liu, Luo, Shao, Xia, Sen, Ratner,
  Hancock, Alborzi, Kuchhal, R{\'{e}}, and
  Malkin}]{distant/bach19snorkeldrybell}
Stephen~H. Bach, Daniel Rodriguez, Yintao Liu, Chong Luo, Haidong Shao,
  Cassandra Xia, Souvik Sen, Alexander Ratner, Braden Hancock, Houman Alborzi,
  Rahul Kuchhal, Christopher R{\'{e}}, and Rob Malkin. 2019.
\newblock \href {https://doi.org/10.1145/3299869.3314036} {Snorkel drybell: {A}
  case study in deploying weak supervision at industrial scale}.
\newblock In \emph{Proceedings of the 2019 International Conference on
  Management of Data, {SIGMOD} Conference 2019, Amsterdam, The Netherlands,
  June 30 - July 5, 2019}, pages 362--375. {ACM}.

\bibitem[{{Banik} et~al.(2019){Banik}, {Hafizur Rahman}, {Chakraborty},
  {Seddiqui}, and {Azim}}]{survey/banik2019bengali}
N.~{Banik}, M.~H. {Hafizur Rahman}, S.~{Chakraborty}, H.~{Seddiqui}, and M.~A.
  {Azim}. 2019.
\newblock \href
  {https://www.researchgate.net/publication/338072328_Survey_on_Text-Based_Sentiment_Analysis_of_Bengali_Language}
  {Survey on text-based sentiment analysis of bengali language}.
\newblock In \emph{2019 1st International Conference on Advances in Science,
  Engineering and Robotics Technology (ICASERT)}, pages 1--6.

\bibitem[{Bansal et~al.(2020)Bansal, Jha, Munkhdalai, and
  McCallum}]{bansal-etal-2020-self}
Trapit Bansal, Rishikesh Jha, Tsendsuren Munkhdalai, and Andrew McCallum. 2020.
\newblock \href {https://doi.org/10.18653/v1/2020.emnlp-main.38}
  {Self-supervised meta-learning for few-shot natural language classification
  tasks}.
\newblock In \emph{Proceedings of the 2020 Conference on Empirical Methods in
  Natural Language Processing (EMNLP)}, pages 522--534, Online. Association for
  Computational Linguistics.

\bibitem[{Bashir et~al.(2017)Bashir, Rozaimee, and
  Isa}]{app/bashir2017automatic}
Muazzam Bashir, Azilawati Rozaimee, and Wan Malini~Wan Isa. 2017.
\newblock \href
  {https://www.researchgate.net/publication/322152904_Automatic_Hausa_LanguageText_Summarization_Based_on_Feature_Extraction_using_Naive_Bayes_Model}
  {Automatic hausa language text summarization based on feature extraction
  using naive bayes model}.
\newblock \emph{World Applied Science Journal}, 35(9):2074--2080.

\bibitem[{Beltagy et~al.(2019)Beltagy, Lo, and
  Cohan}]{domain/beltagy2019scibert}
Iz~Beltagy, Kyle Lo, and Arman Cohan. 2019.
\newblock \href {https://doi.org/10.18653/v1/D19-1371} {{S}ci{BERT}: A
  pretrained language model for scientific text}.
\newblock In \emph{Proceedings of the 2019 Conference on Empirical Methods in
  Natural Language Processing and the 9th International Joint Conference on
  Natural Language Processing (EMNLP-IJCNLP)}, pages 3615--3620, Hong Kong,
  China. Association for Computational Linguistics.

\bibitem[{Bender(2019)}]{intro/bender2019rule}
Emily Bender. 2019.
\newblock \href
  {https://thegradient.pub/the-benderrule-on-naming-the-languages-we-study-and-why-it-matters/}
  {The benderrule: On naming the languages we study and why it matters}.
\newblock \emph{The Gradient}.

\bibitem[{Bhattacharjee et~al.(2020)Bhattacharjee, Ballesteros, Anubhai,
  Muresan, Ma, Ladhak, and Al-Onaizan}]{bert/bhattacharjee-etal-2020-bert}
Kasturi Bhattacharjee, Miguel Ballesteros, Rishita Anubhai, Smaranda Muresan,
  Jie Ma, Faisal Ladhak, and Yaser Al-Onaizan. 2020.
\newblock \href {https://doi.org/10.18653/v1/2020.emnlp-main.636} {To {BERT} or
  not to {BERT}: Comparing task-specific and task-agnostic semi-supervised
  approaches for sequence tagging}.
\newblock In \emph{Proceedings of the 2020 Conference on Empirical Methods in
  Natural Language Processing (EMNLP)}, pages 7927--7934, Online. Association
  for Computational Linguistics.

\bibitem[{Biljon et~al.(2020)Biljon, Pretorius, and
  Kreutzer}]{pre-train/van2020optimal}
Elan~Van Biljon, Arnu Pretorius, and Julia Kreutzer. 2020.
\newblock \href {https://arxiv.org/abs/2004.04418} {On optimal transformer
  depth for low-resource language translation.}
\newblock \emph{CoRR}, abs/2004.04418.

\bibitem[{Bojanowski et~al.(2017)Bojanowski, Grave, Joulin, and
  Mikolov}]{emb/bojanowski2017enriching}
Piotr Bojanowski, Edouard Grave, Armand Joulin, and Tomas Mikolov. 2017.
\newblock \href {https://doi.org/10.1162/tacl_a_00051} {Enriching word vectors
  with subword information}.
\newblock \emph{Transactions of the Association for Computational Linguistics},
  5:135--146.

\bibitem[{Bojar and Tamchyna(2011)}]{data-aug/bojar2011backtranslation}
Ond{\v{r}}ej Bojar and Ale{\v{s}} Tamchyna. 2011.
\newblock \href {https://www.aclweb.org/anthology/W11-2138} {Improving
  translation model by monolingual data}.
\newblock In \emph{Proceedings of the Sixth Workshop on Statistical Machine
  Translation}, pages 330--336, Edinburgh, Scotland. Association for
  Computational Linguistics.

\bibitem[{Cao et~al.(2020)Cao, Kitaev, and Klein}]{multi-ling/cao2020align}
Steven Cao, Nikita Kitaev, and Dan Klein. 2020.
\newblock \href {https://openreview.net/forum?id=r1xCMyBtPS} {Multilingual
  alignment of contextual word representations}.
\newblock In \emph{International Conference on Learning Representations}.

\bibitem[{Cao et~al.(2019)Cao, Hu, Chua, Liu, and Ji}]{distant/cao2019low}
Yixin Cao, Zikun Hu, Tat-seng Chua, Zhiyuan Liu, and Heng Ji. 2019.
\newblock \href {https://doi.org/10.18653/v1/D19-1025} {Low-resource name
  tagging learned with weakly labeled data}.
\newblock In \emph{Proceedings of the 2019 Conference on Empirical Methods in
  Natural Language Processing and the 9th International Joint Conference on
  Natural Language Processing (EMNLP-IJCNLP)}, pages 261--270, Hong Kong,
  China. Association for Computational Linguistics.

\bibitem[{Chen et~al.(2020{\natexlab{a}})Chen, Li, Lei, and
  Shen}]{distant/chen-etal-2020-relabel-agents}
Daoyuan Chen, Yaliang Li, Kai Lei, and Ying Shen. 2020{\natexlab{a}}.
\newblock \href {https://www.aclweb.org/anthology/2020.acl-main.527} {Relabel
  the noise: Joint extraction of entities and relations via cooperative
  multiagents}.
\newblock In \emph{Proceedings of the 58th Annual Meeting of the Association
  for Computational Linguistics}, pages 5940--5950, Online. Association for
  Computational Linguistics.

\bibitem[{Chen et~al.(2019)Chen, Zhang, Mao, Guo, and
  Xu}]{distant/chen-etal-2019-uncover}
Junfan Chen, Richong Zhang, Yongyi Mao, Hongyu Guo, and Jie Xu. 2019.
\newblock \href {https://doi.org/10.18653/v1/D19-1031} {Uncover the
  ground-truth relations in distant supervision: A neural
  expectation-maximization framework}.
\newblock In \emph{Proceedings of the 2019 Conference on Empirical Methods in
  Natural Language Processing and the 9th International Joint Conference on
  Natural Language Processing (EMNLP-IJCNLP)}, pages 326--336, Hong Kong,
  China. Association for Computational Linguistics.

\bibitem[{Chen et~al.(2020{\natexlab{b}})Chen, Ghoshal, Mehdad, Zettlemoyer,
  and Gupta}]{transfer/chen2020domain}
Xilun Chen, Asish Ghoshal, Yashar Mehdad, Luke Zettlemoyer, and Sonal Gupta.
  2020{\natexlab{b}}.
\newblock \href {https://doi.org/10.18653/v1/2020.emnlp-main.413} {Low-resource
  domain adaptation for compositional task-oriented semantic parsing}.
\newblock In \emph{Proceedings of the 2020 Conference on Empirical Methods in
  Natural Language Processing (EMNLP)}, pages 5090--5100, Online. Association
  for Computational Linguistics.

\bibitem[{Chen et~al.(2018)Chen, Sun, Athiwaratkun, Cardie, and
  Weinberger}]{domain/chen2018adversarial}
Xilun Chen, Yu~Sun, Ben Athiwaratkun, Claire Cardie, and Kilian Weinberger.
  2018.
\newblock \href {https://doi.org/10.1162/tacl_a_00039} {Adversarial deep
  averaging networks for cross-lingual sentiment classification}.
\newblock \emph{Transactions of the Association for Computational Linguistics},
  6:557--570.

\bibitem[{Christodoulopoulos and
  Steedman(2015)}]{projections/Christodoulopoulos15Bible}
Christos Christodoulopoulos and Mark Steedman. 2015.
\newblock \href {https://doi.org/10.1007/s10579-014-9287-y} {A massively
  parallel corpus: the bible in 100 languages}.
\newblock \emph{Lang. Resour. Evaluation}, 49(2):375--395.

\bibitem[{Cieri et~al.(2016)Cieri, Maxwell, Strassel, and
  Tracey}]{survey/cieri2016selection}
Christopher Cieri, Mike Maxwell, Stephanie Strassel, and Jennifer Tracey. 2016.
\newblock \href {https://www.aclweb.org/anthology/L16-1720} {Selection criteria
  for low resource language programs}.
\newblock In \emph{Proceedings of the Tenth International Conference on
  Language Resources and Evaluation ({LREC}'16)}, pages 4543--4549,
  Portoro{\v{z}}, Slovenia. European Language Resources Association (ELRA).

\bibitem[{Clark et~al.(2018)Clark, Luong, Manning, and
  Le}]{cvt/clark-etal-2018-semi}
Kevin Clark, Minh-Thang Luong, Christopher~D. Manning, and Quoc Le. 2018.
\newblock \href {https://doi.org/10.18653/v1/D18-1217} {Semi-supervised
  sequence modeling with cross-view training}.
\newblock In \emph{Proceedings of the 2018 Conference on Empirical Methods in
  Natural Language Processing}, pages 1914--1925, Brussels, Belgium.
  Association for Computational Linguistics.

\bibitem[{Collobert et~al.(2011)Collobert, Weston, Bottou, Karlen, Kavukcuoglu,
  and Kuksa}]{pre-train/collobert2011natural}
Ronan Collobert, Jason Weston, L{\'e}on Bottou, Michael Karlen, Koray
  Kavukcuoglu, and Pavel Kuksa. 2011.
\newblock \href
  {https://jmlr.org/papers/volume12/collobert11a/collobert11a.pdf} {Natural
  language processing (almost) from scratch}.
\newblock \emph{Journal of machine learning research}, 12(ARTICLE):2493--2537.

\bibitem[{Conneau et~al.(2020)Conneau, Khandelwal, Goyal, Chaudhary, Wenzek,
  Guzm{\'a}n, Grave, Ott, Zettlemoyer, and
  Stoyanov}]{pre-train/conneau2020unsupervised}
Alexis Conneau, Kartikay Khandelwal, Naman Goyal, Vishrav Chaudhary, Guillaume
  Wenzek, Francisco Guzm{\'a}n, Edouard Grave, Myle Ott, Luke Zettlemoyer, and
  Veselin Stoyanov. 2020.
\newblock \href {https://www.aclweb.org/anthology/2020.acl-main.747}
  {Unsupervised cross-lingual representation learning at scale}.
\newblock In \emph{Proceedings of the 58th Annual Meeting of the Association
  for Computational Linguistics}, pages 8440--8451, Online. Association for
  Computational Linguistics.

\bibitem[{Cotterell et~al.(2018)Cotterell, Kirov, Sylak-Glassman, Walther,
  Vylomova, McCarthy, Kann, Mielke, Nicolai, Silfverberg, Yarowsky, Eisner, and
  Hulden}]{survey/conll2018universalMorphologicalReinflection}
Ryan Cotterell, Christo Kirov, John Sylak-Glassman, G{\'e}raldine Walther,
  Ekaterina Vylomova, Arya~D. McCarthy, Katharina Kann, Sabrina~J. Mielke,
  Garrett Nicolai, Miikka Silfverberg, David Yarowsky, Jason Eisner, and Mans
  Hulden. 2018.
\newblock \href {https://doi.org/10.18653/v1/K18-3001} {The
  {C}o{NLL}{--}{SIGMORPHON} 2018 shared task: Universal morphological
  reinflection}.
\newblock In \emph{Proceedings of the {C}o{NLL}{--}{SIGMORPHON} 2018 Shared
  Task: Universal Morphological Reinflection}, pages 1--27, Brussels.
  Association for Computational Linguistics.

\bibitem[{Cruz and Cheng(2019)}]{pre-train/cruz2019evaluating}
Jan Christian~Blaise Cruz and Charibeth Cheng. 2019.
\newblock \href {https://arxiv.org/abs/1907.00409} {Evaluating language model
  finetuning techniques for low-resource languages}.
\newblock \emph{arXiv preprint arXiv:1907.00409}.

\bibitem[{Dai and Adel(2020)}]{data-aug/dai2020analysis}
Xiang Dai and Heike Adel. 2020.
\newblock \href {https://doi.org/10.18653/v1/2020.coling-main.343} {An analysis
  of simple data augmentation for named entity recognition}.
\newblock In \emph{Proceedings of the 28th International Conference on
  Computational Linguistics}, pages 3861--3867, Barcelona, Spain (Online).
  International Committee on Computational Linguistics.

\bibitem[{Daud et~al.(2017)Daud, Khan, and Che}]{survey/daud2017urdu}
Ali Daud, Wahab Khan, and Dunren Che. 2017.
\newblock \href
  {https://link.springer.com/content/pdf/10.1007/s10462-016-9482-x.pdf} {Urdu
  language processing: a survey}.
\newblock \emph{Artificial Intelligence Review}, 47(3):279--311.

\bibitem[{De~Pauw et~al.(2011)De~Pauw, De~Schryver, Pretorius, and
  Levin}]{survey/de2011introduction}
Guy De~Pauw, Gilles-Maurice De~Schryver, Laurette Pretorius, and Lori Levin.
  2011.
\newblock \href
  {https://d1wqtxts1xzle7.cloudfront.net/34611293/LRE_45.3_Intro.pdf?1409706958=&response-content-disposition=inline\%3B+filename\%3DIntroduction_to_the_special_issue_on_Afr.pdf&Expires=1593092515&Signature=RifwTvamVqy-VjIvmc63NO5kWovlZB-7Pk73kPNcARVLxLlHl1uooN2O0T6vtuhH2zVsMvot7k4vdLO7W8SsJ0vACZPe8y9hdnyaQPhLcJfllDBtw7ebdQ2jka84XRzmeh6RDpSn4E0hyU599t~zpY0j6GWO0JhRHktomQpBKTKwCiiX7-4b-flXxElxyqaJlZuRlDZMEvT0xEuW77VvwFe4vLVtzofgZUIppV4evc0EaFCfAvPof~ixZMGNx7RJKyfL2P7gGPQHIdR6hNd~HKTU6e1izXpivceDTfU64pxIGDWZ0e6q5ZFjnD5Y~2YfUSQuaAolCdZiikou2Q1uWQ__&Key-Pair-Id=APKAJLOHF5GGSLRBV4ZA}
  {Introduction to the special issue on african language technology}.
\newblock \emph{Language Resources and Evaluation}, 45(3):263--269.

\bibitem[{Dehouck and G{\'o}mez-Rodr{\'\i}guez(2020)}]{dehouck2020swapping}
Mathieu Dehouck and Carlos G{\'o}mez-Rodr{\'\i}guez. 2020.
\newblock \href {https://doi.org/10.18653/v1/2020.coling-main.339} {Data
  augmentation via subtree swapping for dependency parsing of low-resource
  languages}.
\newblock In \emph{Proceedings of the 28th International Conference on
  Computational Linguistics}, pages 3818--3830, Barcelona, Spain (Online).
  International Committee on Computational Linguistics.

\bibitem[{Deng and Sun(2019)}]{distant/deng2019hop2}
Xiang Deng and Huan Sun. 2019.
\newblock \href {https://doi.org/10.18653/v1/D19-1039} {Leveraging 2-hop
  distant supervision from table entity pairs for relation extraction}.
\newblock In \emph{Proceedings of the 2019 Conference on Empirical Methods in
  Natural Language Processing and the 9th International Joint Conference on
  Natural Language Processing (EMNLP-IJCNLP)}, pages 410--420, Hong Kong,
  China. Association for Computational Linguistics.

\bibitem[{Devlin et~al.(2019)Devlin, Chang, Lee, and
  Toutanova}]{pre-train/devlin2019bert}
Jacob Devlin, Ming-Wei Chang, Kenton Lee, and Kristina Toutanova. 2019.
\newblock \href {https://doi.org/10.18653/v1/N19-1423} {{BERT}: Pre-training of
  deep bidirectional transformers for language understanding}.
\newblock In \emph{Proceedings of the 2019 Conference of the North {A}merican
  Chapter of the Association for Computational Linguistics: Human Language
  Technologies, Volume 1 (Long and Short Papers)}, pages 4171--4186,
  Minneapolis, Minnesota. Association for Computational Linguistics.

\bibitem[{Ding et~al.(2020)Ding, Liu, Bing, Kruengkrai, Nguyen, Joty, Si, and
  Miao}]{data-aug/ding2020daga}
Bosheng Ding, Linlin Liu, Lidong Bing, Canasai Kruengkrai, Thien~Hai Nguyen,
  Shafiq Joty, Luo Si, and Chunyan Miao. 2020.
\newblock \href {https://doi.org/10.18653/v1/2020.emnlp-main.488} {{DAGA}: Data
  augmentation with a generation approach for low-resource tagging tasks}.
\newblock In \emph{Proceedings of the 2020 Conference on Empirical Methods in
  Natural Language Processing (EMNLP)}, pages 6045--6057, Online. Association
  for Computational Linguistics.

\bibitem[{Dou et~al.(2019)Dou, Yu, and Anastasopoulos}]{transfer/dou2019meta}
Zi-Yi Dou, Keyi Yu, and Antonios Anastasopoulos. 2019.
\newblock \href {https://doi.org/10.18653/v1/D19-1112} {Investigating
  meta-learning algorithms for low-resource natural language understanding
  tasks}.
\newblock In \emph{Proceedings of the 2019 Conference on Empirical Methods in
  Natural Language Processing and the 9th International Joint Conference on
  Natural Language Processing (EMNLP-IJCNLP)}, pages 1192--1197, Hong Kong,
  China. Association for Computational Linguistics.

\bibitem[{Eberhard et~al.(2019)Eberhard, Simons, and
  (eds.)}]{intro/Ethnologue2019}
David~M. Eberhard, Gary~F. Simons, and Charles D.~Fennig (eds.). 2019.
\newblock \href {http://www.ethnologue.com} {Ethnologue: Languages of the
  world. twenty-second edition.}

\bibitem[{Eskander et~al.(2020)Eskander, Muresan, and
  Collins}]{projection/eskander2020unsupervised}
Ramy Eskander, Smaranda Muresan, and Michael Collins. 2020.
\newblock \href {https://doi.org/10.18653/v1/2020.emnlp-main.391} {Unsupervised
  cross-lingual part-of-speech tagging for truly low-resource scenarios}.
\newblock In \emph{Proceedings of the 2020 Conference on Empirical Methods in
  Natural Language Processing (EMNLP)}, pages 4820--4831, Online. Association
  for Computational Linguistics.

\bibitem[{Fabuni and Salawu(2005)}]{app/fabuni2005yoruba}
Felix~Abidemi Fabuni and Akeem~Segun Salawu. 2005.
\newblock \href {https://www.njas.fi/njas/article/view/262} {Is yor{\`u}b{\'a}
  an endangered language?}
\newblock \emph{Nordic Journal of African Studies}, 14(3):18--18.

\bibitem[{Fadaee et~al.(2017)Fadaee, Bisazza, and
  Monz}]{data-aug/fadaee-etal-2017-data}
Marzieh Fadaee, Arianna Bisazza, and Christof Monz. 2017.
\newblock \href {https://doi.org/10.18653/v1/P17-2090} {Data augmentation for
  low-resource neural machine translation}.
\newblock In \emph{Proceedings of the 55th Annual Meeting of the Association
  for Computational Linguistics (Volume 2: Short Papers)}, pages 567--573,
  Vancouver, Canada. Association for Computational Linguistics.

\bibitem[{Fang and Cohn(2016)}]{distant/fang-cohn-2016-learning}
Meng Fang and Trevor Cohn. 2016.
\newblock \href {https://doi.org/10.18653/v1/K16-1018} {Learning when to trust
  distant supervision: An application to low-resource {POS} tagging using
  cross-lingual projection}.
\newblock In \emph{Proceedings of The 20th {SIGNLL} Conference on Computational
  Natural Language Learning}, pages 178--186, Berlin, Germany. Association for
  Computational Linguistics.

\bibitem[{Fang and Cohn(2017)}]{projection/fang-cohn-2017-model}
Meng Fang and Trevor Cohn. 2017.
\newblock \href {https://doi.org/10.18653/v1/P17-2093} {Model transfer for
  tagging low-resource languages using a bilingual dictionary}.
\newblock In \emph{Proceedings of the 55th Annual Meeting of the Association
  for Computational Linguistics (Volume 2: Short Papers)}, pages 587--593,
  Vancouver, Canada. Association for Computational Linguistics.

\bibitem[{Fei et~al.(2020)Fei, Zhang, and Ji}]{projection/fei-etal-2020-cross}
Hao Fei, Meishan Zhang, and Donghong Ji. 2020.
\newblock \href {https://www.aclweb.org/anthology/2020.acl-main.627}
  {Cross-lingual semantic role labeling with high-quality translated training
  corpus}.
\newblock In \emph{Proceedings of the 58th Annual Meeting of the Association
  for Computational Linguistics}, pages 7014--7026, Online. Association for
  Computational Linguistics.

\bibitem[{Finn et~al.(2017)Finn, Abbeel, and Levine}]{transfer/Finn17Meta}
Chelsea Finn, Pieter Abbeel, and Sergey Levine. 2017.
\newblock \href {http://proceedings.mlr.press/v70/finn17a} {Model-agnostic
  meta-learning for fast adaptation of deep networks}.
\newblock In \emph{Proceedings of the 34th International Conference on Machine
  Learning - Volume 70}, ICML’17, page 1126–1135. JMLR.org.

\bibitem[{Fr{\'e}nay and Verleysen(2013)}]{survey/frenay2013classification}
Beno{\^\i}t Fr{\'e}nay and Michel Verleysen. 2013.
\newblock \href {https://ieeexplore.ieee.org/abstract/document/6685834/}
  {Classification in the presence of label noise: a survey}.
\newblock \emph{IEEE transactions on neural networks and learning systems},
  25(5):845--869.

\bibitem[{Friedrich et~al.(2020)Friedrich, Adel, Tomazic, Hingerl, Benteau,
  Marusczyk, and Lange}]{domain/friedrich2020sofc}
Annemarie Friedrich, Heike Adel, Federico Tomazic, Johannes Hingerl, Renou
  Benteau, Anika Marusczyk, and Lukas Lange. 2020.
\newblock \href {https://www.aclweb.org/anthology/2020.acl-main.116} {The
  {SOFC}-exp corpus and neural approaches to information extraction in the
  materials science domain}.
\newblock In \emph{Proceedings of the 58th Annual Meeting of the Association
  for Computational Linguistics}, pages 1255--1268, Online. Association for
  Computational Linguistics.

\bibitem[{Friedrich and
  Gateva(2017)}]{projection/friedrich-gateva-2017telicity}
Annemarie Friedrich and Damyana Gateva. 2017.
\newblock \href {https://doi.org/10.18653/v1/D17-1271} {Classification of
  telicity using cross-linguistic annotation projection}.
\newblock In \emph{Proceedings of the 2017 Conference on Empirical Methods in
  Natural Language Processing}, pages 2559--2565, Copenhagen, Denmark.
  Association for Computational Linguistics.

\bibitem[{Garg and Ramakrishnan(2020)}]{adverserial/garg2020BAE}
Siddhant Garg and Goutham Ramakrishnan. 2020.
\newblock \href {https://doi.org/10.18653/v1/2020.emnlp-main.498} {{BAE}:
  {BERT}-based adversarial examples for text classification}.
\newblock In \emph{Proceedings of the 2020 Conference on Empirical Methods in
  Natural Language Processing (EMNLP)}, pages 6174--6181, Online. Association
  for Computational Linguistics.

\bibitem[{Garrette and Baldridge(2013)}]{distant/garrette2013pos2hours}
Dan Garrette and Jason Baldridge. 2013.
\newblock \href {https://www.aclweb.org/anthology/N13-1014} {Learning a
  part-of-speech tagger from two hours of annotation}.
\newblock In \emph{Proceedings of the 2013 Conference of the North {A}merican
  Chapter of the Association for Computational Linguistics: Human Language
  Technologies}, pages 138--147, Atlanta, Georgia. Association for
  Computational Linguistics.

\bibitem[{Goodfellow et~al.(2014)Goodfellow, Pouget-Abadie, Mirza, Xu,
  Warde-Farley, Ozair, Courville, and Bengio}]{transfer/goodfellow2014gan}
Ian Goodfellow, Jean Pouget-Abadie, Mehdi Mirza, Bing Xu, David Warde-Farley,
  Sherjil Ozair, Aaron Courville, and Yoshua Bengio. 2014.
\newblock \href
  {http://papers.nips.cc/paper/5423-generative-adversarial-nets.pdf}
  {Generative adversarial nets}.
\newblock In Z.~Ghahramani, M.~Welling, C.~Cortes, N.~D. Lawrence, and K.~Q.
  Weinberger, editors, \emph{Advances in Neural Information Processing Systems
  27}, pages 2672--2680. Curran Associates, Inc.

\bibitem[{Grie{\ss}haber et~al.(2020)Grie{\ss}haber, Vu, and
  Maucher}]{domain/griesshaber2020low}
Daniel Grie{\ss}haber, Ngoc~Thang Vu, and Johannes Maucher. 2020.
\newblock \href
  {https://www.sciencedirect.com/science/article/pii/S0885230819303006}
  {Low-resource text classification using domain-adversarial learning}.
\newblock \emph{Computer Speech \& Language}, 62:101056.

\bibitem[{Grover et~al.(2010)Grover, Calteaux, van Huyssteen, and
  Pretorius}]{survey/grover2010overview}
Aditi~Sharma Grover, Karen Calteaux, Gerhard van Huyssteen, and Marthinus
  Pretorius. 2010.
\newblock \href {https://dl.acm.org/doi/pdf/10.1145/1899503.1899547} {An
  overview of hlts for south african bantu languages}.
\newblock In \emph{Proceedings of the 2010 Annual Research Conference of the
  South African Institute of Computer Scientists and Information
  Technologists}, pages 370--375.

\bibitem[{Grundkiewicz et~al.(2019)Grundkiewicz, Junczys-Dowmunt, and
  Heafield}]{data-aug/grundkiewicz2019Grammar}
Roman Grundkiewicz, Marcin Junczys-Dowmunt, and Kenneth Heafield. 2019.
\newblock \href {https://doi.org/10.18653/v1/W19-4427} {Neural grammatical
  error correction systems with unsupervised pre-training on synthetic data}.
\newblock In \emph{Proceedings of the Fourteenth Workshop on Innovative Use of
  NLP for Building Educational Applications}, pages 252--263, Florence, Italy.
  Association for Computational Linguistics.

\bibitem[{Guellil et~al.(2019)Guellil, Azouaou, and
  Valitutti}]{survey/guellil2019arabic}
Imane Guellil, Fai{\c{c}}al Azouaou, and Alessandro Valitutti. 2019.
\newblock \href {https://ieeexplore.ieee.org/abstract/document/9035299}
  {English vs arabic sentiment analysis: A survey presenting 100 work studies,
  resources and tools}.
\newblock In \emph{2019 IEEE/ACS 16th International Conference on Computer
  Systems and Applications (AICCSA)}, pages 1--8.

\bibitem[{Gui et~al.(2017)Gui, Zhang, Huang, Peng, and
  Huang}]{transfer/gui-etal-2017-part-adversarial}
Tao Gui, Qi~Zhang, Haoran Huang, Minlong Peng, and Xuanjing Huang. 2017.
\newblock \href {https://doi.org/10.18653/v1/D17-1256} {Part-of-speech tagging
  for twitter with adversarial neural networks}.
\newblock In \emph{Proceedings of the 2017 Conference on Empirical Methods in
  Natural Language Processing}, pages 2411--2420, Copenhagen, Denmark.
  Association for Computational Linguistics.

\bibitem[{Gulordava et~al.(2018)Gulordava, Bojanowski, Grave, Linzen, and
  Baroni}]{data-aug/gulordava2018Colorless}
Kristina Gulordava, Piotr Bojanowski, Edouard Grave, Tal Linzen, and Marco
  Baroni. 2018.
\newblock \href {https://doi.org/10.18653/v1/N18-1108} {Colorless green
  recurrent networks dream hierarchically}.
\newblock In \emph{Proceedings of the 2018 Conference of the North {A}merican
  Chapter of the Association for Computational Linguistics: Human Language
  Technologies, Volume 1 (Long Papers)}, pages 1195--1205, New Orleans,
  Louisiana. Association for Computational Linguistics.

\bibitem[{Gururangan et~al.(2020)Gururangan, Marasovi{\'c}, Swayamdipta, Lo,
  Beltagy, Downey, and Smith}]{domain/gugurangan2020dontstop}
Suchin Gururangan, Ana Marasovi{\'c}, Swabha Swayamdipta, Kyle Lo, Iz~Beltagy,
  Doug Downey, and Noah~A. Smith. 2020.
\newblock \href {https://www.aclweb.org/anthology/2020.acl-main.740} {Don{'}t
  stop pretraining: Adapt language models to domains and tasks}.
\newblock In \emph{Proceedings of the 58th Annual Meeting of the Association
  for Computational Linguistics}, pages 8342--8360, Online. Association for
  Computational Linguistics.

\bibitem[{Hakro et~al.(2016)Hakro, TALIB, and
  Mojai}]{app/hakro2016multilingual}
DN~Hakro, AZ~TALIB, and GN~Mojai. 2016.
\newblock \href
  {https://sujo-old.usindh.edu.pk/index.php/SURJ/article/view/2299}
  {Multilingual text image database for ocr}.
\newblock \emph{Sindh University Research Journal-SURJ (Science Series)},
  47(1).

\bibitem[{Harish and Rangan(2020)}]{survey/harish2020comprehensive}
BS~Harish and R~Kasturi Rangan. 2020.
\newblock \href {https://link.springer.com/article/10.1007/s42452-020-2983-x}
  {A comprehensive survey on indian regional language processing}.
\newblock \emph{SN Applied Sciences}, 2(7):1--16.

\bibitem[{Hedderich et~al.(2020)Hedderich, Adelani, Zhu, Alabi, Markus, and
  Klakow}]{hedderich2020transfer}
Michael~A. Hedderich, David Adelani, Dawei Zhu, Jesujoba Alabi, Udia Markus,
  and Dietrich Klakow. 2020.
\newblock \href {https://doi.org/10.18653/v1/2020.emnlp-main.204} {Transfer
  learning and distant supervision for multilingual transformer models: A study
  on {A}frican languages}.
\newblock In \emph{Proceedings of the 2020 Conference on Empirical Methods in
  Natural Language Processing (EMNLP)}, pages 2580--2591, Online. Association
  for Computational Linguistics.

\bibitem[{Hedderich and Klakow(2018)}]{distant/hedderich2018Noisy}
Michael~A. Hedderich and Dietrich Klakow. 2018.
\newblock \href {https://doi.org/10.18653/v1/W18-3402} {Training a neural
  network in a low-resource setting on automatically annotated noisy data}.
\newblock In \emph{Proceedings of the Workshop on Deep Learning Approaches for
  Low-Resource {NLP}}, pages 12--18, Melbourne. Association for Computational
  Linguistics.

\bibitem[{Hedderich et~al.(2021{\natexlab{a}})Hedderich, Lange, and
  Klakow}]{distant/hedderich2021ANEA}
Michael~A. Hedderich, Lukas Lange, and Dietrich Klakow. 2021{\natexlab{a}}.
\newblock \href {http://arxiv.org/abs/2102.13129} {{ANEA:} distant supervision
  for low-resource named entity recognition}.
\newblock \emph{CoRR}, abs/2102.13129.

\bibitem[{Hedderich et~al.(2021{\natexlab{b}})Hedderich, Zhu, and
  Klakow}]{distant/hedderich21noisemodel}
Michael~A. Hedderich, Dawei Zhu, and Dietrich Klakow. 2021{\natexlab{b}}.
\newblock \href {https://arxiv.org/abs/2101.09763} {Analysing the noise model
  error for realistic noisy label data}.
\newblock In \emph{Proceedings of AAAI}.

\bibitem[{Hegde and Patil(2020)}]{data-aug/hegde2020paraphrase}
Chaitra Hegde and Shrikumar Patil. 2020.
\newblock \href {http://arxiv.org/abs/2006.05477} {Unsupervised paraphrase
  generation using pre-trained language models}.
\newblock \emph{CoRR}, abs/2006.05477.

\bibitem[{Heinzerling and Strube(2018)}]{emb/heinzerling2018bpemb}
Benjamin Heinzerling and Michael Strube. 2018.
\newblock \href {https://www.aclweb.org/anthology/L18-1473} {{BPEmb:
  Tokenization-free Pre-trained Subword Embeddings in 275 Languages}}.
\newblock In \emph{Proceedings of the Eleventh International Conference on
  Language Resources and Evaluation (LREC 2018)}, Miyazaki, Japan. European
  Language Resources Association (ELRA).

\bibitem[{Hoang et~al.(2018)Hoang, Koehn, Haffari, and
  Cohn}]{data-aug/hoang2018iterativeBacktranslation}
Vu~Cong~Duy Hoang, Philipp Koehn, Gholamreza Haffari, and Trevor Cohn. 2018.
\newblock \href {https://doi.org/10.18653/v1/W18-2703} {Iterative
  back-translation for neural machine translation}.
\newblock In \emph{Proceedings of the 2nd Workshop on Neural Machine
  Translation and Generation}, pages 18--24, Melbourne, Australia. Association
  for Computational Linguistics.

\bibitem[{Hospedales et~al.(2020)Hospedales, Antoniou, Micaelli, and
  Storkey}]{survey/hospedales2020meta}
Timothy Hospedales, Antreas Antoniou, Paul Micaelli, and Amos Storkey. 2020.
\newblock \href {https://arxiv.org/abs/2004.05439} {Meta-learning in neural
  networks: A survey}.
\newblock \emph{arXiv preprint arXiv:2004.05439}.

\bibitem[{Hsu et~al.(2019)Hsu, Liu, and Lee}]{cross-ling/hsu-etal-2019-zero}
Tsung-Yuan Hsu, Chi-Liang Liu, and Hung-yi Lee. 2019.
\newblock \href {https://doi.org/10.18653/v1/D19-1607} {Zero-shot reading
  comprehension by cross-lingual transfer learning with multi-lingual language
  representation model}.
\newblock In \emph{Proceedings of the 2019 Conference on Empirical Methods in
  Natural Language Processing and the 9th International Joint Conference on
  Natural Language Processing (EMNLP-IJCNLP)}, pages 5933--5940, Hong Kong,
  China. Association for Computational Linguistics.

\bibitem[{Hu et~al.(2020)Hu, Ruder, Siddhant, Neubig, Firat, and
  Johnson}]{eval/hu2020xtreme}
Junjie Hu, Sebastian Ruder, Aditya Siddhant, Graham Neubig, Orhan Firat, and
  Melvin Johnson. 2020.
\newblock \href {http://proceedings.mlr.press/v119/hu20b.html} {Xtreme: A
  massively multilingual multi-task benchmark for evaluating cross-lingual
  generalisation}.
\newblock \emph{International Conference on Machine Learning}, pages
  4411--4421.

\bibitem[{Hu et~al.(2019)Hu, Zhang, Shi, Nie, Guan, and
  Yang}]{distant/hu2019jointLabel}
Linmei Hu, Luhao Zhang, Chuan Shi, Liqiang Nie, Weili Guan, and Cheng Yang.
  2019.
\newblock \href {https://doi.org/10.18653/v1/D19-1395} {Improving
  distantly-supervised relation extraction with joint label embedding}.
\newblock In \emph{Proceedings of the 2019 Conference on Empirical Methods in
  Natural Language Processing and the 9th International Joint Conference on
  Natural Language Processing (EMNLP-IJCNLP)}, pages 3821--3829, Hong Kong,
  China. Association for Computational Linguistics.

\bibitem[{Huang et~al.(2019)Huang, Altosaar, and
  Ranganath}]{domain/huang2019clinicalbert}
Kexin Huang, Jaan Altosaar, and Rajesh Ranganath. 2019.
\newblock \href {https://arxiv.org/abs/1904.05342} {Clinicalbert: Modeling
  clinical notes and predicting hospital readmission}.
\newblock \emph{arXiv preprint arXiv:1904.05342}.

\bibitem[{Huang et~al.(2020)Huang, Feng, Ma, Du, and
  Wu}]{transfer/huang2020meta}
Yi~Huang, Junlan Feng, Shuo Ma, Xiaoyu Du, and Xiaoting Wu. 2020.
\newblock \href {https://doi.org/10.18653/v1/2020.findings-emnlp.368} {Towards
  low-resource semi-supervised dialogue generation with meta-learning}.
\newblock In \emph{Findings of the Association for Computational Linguistics:
  EMNLP 2020}, pages 4123--4128, Online. Association for Computational
  Linguistics.

\bibitem[{Huang and Du(2019)}]{distant/huang2019curriculum}
Yuyun Huang and Jinhua Du. 2019.
\newblock \href {https://doi.org/10.18653/v1/D19-1037} {Self-attention enhanced
  {CNN}s and collaborative curriculum learning for distantly supervised
  relation extraction}.
\newblock In \emph{Proceedings of the 2019 Conference on Empirical Methods in
  Natural Language Processing and the 9th International Joint Conference on
  Natural Language Processing (EMNLP-IJCNLP)}, pages 389--398, Hong Kong,
  China. Association for Computational Linguistics.

\bibitem[{Huber and Carenini(2020)}]{huber2020discourse}
Patrick Huber and Giuseppe Carenini. 2020.
\newblock \href {https://doi.org/10.18653/v1/2020.emnlp-main.603} {{MEGA} {RST}
  discourse treebanks with structure and nuclearity from scalable distant
  sentiment supervision}.
\newblock In \emph{Proceedings of the 2020 Conference on Empirical Methods in
  Natural Language Processing (EMNLP)}, pages 7442--7457, Online. Association
  for Computational Linguistics.

\bibitem[{Hvingelby et~al.(2020)Hvingelby, Pauli, Barrett, Rosted, Lidegaard,
  and S{\o}gaard}]{cross-ling/hvingelby-etal-2020-dane}
Rasmus Hvingelby, Amalie~Brogaard Pauli, Maria Barrett, Christina Rosted,
  Lasse~Malm Lidegaard, and Anders S{\o}gaard. 2020.
\newblock \href {https://www.aclweb.org/anthology/2020.lrec-1.565} {{D}a{NE}: A
  named entity resource for {D}anish}.
\newblock In \emph{Proceedings of the 12th Language Resources and Evaluation
  Conference}, pages 4597--4604, Marseille, France. European Language Resources
  Association.

\bibitem[{Jain and Ganesamoorty(2020)}]{domain/jain2020nukebert}
Ayush Jain and Meenachi Ganesamoorty. 2020.
\newblock \href {https://arxiv.org/abs/2003.13821} {Nukebert: A pre-trained
  language model for low resource nuclear domain}.
\newblock \emph{arXiv preprint arXiv:2003.13821}.

\bibitem[{Jia et~al.(2019)Jia, Dai, Xiao, and Wu}]{distant/jia2019Arnor}
Wei Jia, Dai Dai, Xinyan Xiao, and Hua Wu. 2019.
\newblock \href {https://doi.org/10.18653/v1/P19-1135} {{ARNOR}: Attention
  regularization based noise reduction for distant supervision relation
  classification}.
\newblock In \emph{Proceedings of the 57th Annual Meeting of the Association
  for Computational Linguistics}, pages 1399--1408, Florence, Italy.
  Association for Computational Linguistics.

\bibitem[{Jie et~al.(2019)Jie, Xie, Lu, Ding, and
  Li}]{distant/jie-etal-2019-betterIncomplete}
Zhanming Jie, Pengjun Xie, Wei Lu, Ruixue Ding, and Linlin Li. 2019.
\newblock \href {https://doi.org/10.18653/v1/N19-1079} {Better modeling of
  incomplete annotations for named entity recognition}.
\newblock In \emph{Proceedings of the 2019 Conference of the North {A}merican
  Chapter of the Association for Computational Linguistics: Human Language
  Technologies, Volume 1 (Long and Short Papers)}, pages 729--734, Minneapolis,
  Minnesota. Association for Computational Linguistics.

\bibitem[{Jin et~al.(2020)Jin, Jin, Zhou, and
  Szolovits}]{adverserial/jin2020BERTrobust}
Di~Jin, Zhijing Jin, Joey~Tianyi Zhou, and Peter Szolovits. 2020.
\newblock \href {https://aaai.org/ojs/index.php/AAAI/article/view/6311} {Is
  {BERT} really robust? {A} strong baseline for natural language attack on text
  classification and entailment}.
\newblock In \emph{The Thirty-Fourth {AAAI} Conference on Artificial
  Intelligence, {AAAI} 2020, The Thirty-Second Innovative Applications of
  Artificial Intelligence Conference, {IAAI} 2020, The Tenth {AAAI} Symposium
  on Educational Advances in Artificial Intelligence, {EAAI} 2020, New York,
  NY, USA, February 7-12, 2020}, pages 8018--8025. {AAAI} Press.

\bibitem[{Joshi et~al.(2020)Joshi, Santy, Budhiraja, Bali, and
  Choudhury}]{survey/joshi2020linguisticdiversity}
Pratik Joshi, Sebastin Santy, Amar Budhiraja, Kalika Bali, and Monojit
  Choudhury. 2020.
\newblock \href {https://doi.org/10.18653/v1/2020.acl-main.560} {The state and
  fate of linguistic diversity and inclusion in the {NLP} world}.
\newblock In \emph{Proceedings of the 58th Annual Meeting of the Association
  for Computational Linguistics}, pages 6282--6293, Online. Association for
  Computational Linguistics.

\bibitem[{Joulin et~al.(2018)Joulin, Bojanowski, Mikolov, J{\'e}gou, and
  Grave}]{multi-ling/joulin2018loss}
Armand Joulin, Piotr Bojanowski, Tomas Mikolov, Herv{\'e} J{\'e}gou, and
  Edouard Grave. 2018.
\newblock \href {https://doi.org/10.18653/v1/D18-1330} {Loss in translation:
  Learning bilingual word mapping with a retrieval criterion}.
\newblock In \emph{Proceedings of the 2018 Conference on Empirical Methods in
  Natural Language Processing}.

\bibitem[{Jungmaier et~al.(2020)Jungmaier, Kassner, and
  Roth}]{emb/jungmaier2020dirichlet}
Jakob Jungmaier, Nora Kassner, and Benjamin Roth. 2020.
\newblock \href {https://www.aclweb.org/anthology/2020.lrec-1.437}
  {{D}irichlet-smoothed word embeddings for low-resource settings}.
\newblock In \emph{Proceedings of The 12th Language Resources and Evaluation
  Conference}, pages 3560--3565, Marseille, France. European Language Resources
  Association.

\bibitem[{Kann et~al.(2020)Kann, Lacroix, and
  S{\o}gaard}]{defining/kann2020POSPoorly}
Katharina Kann, Oph{\'{e}}lie Lacroix, and Anders S{\o}gaard. 2020.
\newblock \href {https://aaai.org/ojs/index.php/AAAI/article/view/6317} {Weakly
  supervised {POS} taggers perform poorly on \emph{Truly} low-resource
  languages}.
\newblock In \emph{The Thirty-Fourth {AAAI} Conference on Artificial
  Intelligence, {AAAI} 2020, The Thirty-Second Innovative Applications of
  Artificial Intelligence Conference, {IAAI} 2020, The Tenth {AAAI} Symposium
  on Educational Advances in Artificial Intelligence, {EAAI} 2020, New York,
  NY, USA, February 7-12, 2020}, pages 8066--8073. {AAAI} Press.

\bibitem[{Karamanolakis et~al.(2019)Karamanolakis, Hsu, and
  Gravano}]{distant/karamanolakis-etal-2019-leveraging}
Giannis Karamanolakis, Daniel Hsu, and Luis Gravano. 2019.
\newblock \href {https://doi.org/10.18653/v1/D19-1468} {Leveraging just a few
  keywords for fine-grained aspect detection through weakly supervised
  co-training}.
\newblock In \emph{Proceedings of the 2019 Conference on Empirical Methods in
  Natural Language Processing and the 9th International Joint Conference on
  Natural Language Processing (EMNLP-IJCNLP)}, pages 4611--4621, Hong Kong,
  China. Association for Computational Linguistics.

\bibitem[{Karamanolakis et~al.(2020)Karamanolakis, Hsu, and
  Gravano}]{cross-ling/karamanolakis2020seedwords}
Giannis Karamanolakis, Daniel Hsu, and Luis Gravano. 2020.
\newblock \href {https://doi.org/10.18653/v1/2020.findings-emnlp.323}
  {Cross-lingual text classification with minimal resources by transferring a
  sparse teacher}.
\newblock In \emph{Findings of the Association for Computational Linguistics:
  EMNLP 2020}, pages 3604--3622, Online. Association for Computational
  Linguistics.

\bibitem[{Karamanolakis et~al.(2021)Karamanolakis, Mukherjee, Zheng, and
  Awadallah}]{karamanolakis2021leaving}
Giannis Karamanolakis, Subhabrata~(Subho) Mukherjee, Guoqing Zheng, and
  Ahmed~H. Awadallah. 2021.
\newblock \href
  {https://www.microsoft.com/en-us/research/publication/leaving-no-valuable-knowledge-behind-weak-supervision-with-self-training-and-domain-specific-rules/}
  {Leaving no valuable knowledge behind: Weak supervision with self-training
  and domain-specific rules}.
\newblock In \emph{NAACL 2021}. NAACL 2021.

\bibitem[{Kasai et~al.(2019)Kasai, Qian, Gurajada, Li, and
  Popa}]{transfer/kasai2019adversarial}
Jungo Kasai, Kun Qian, Sairam Gurajada, Yunyao Li, and Lucian Popa. 2019.
\newblock \href {https://doi.org/10.18653/v1/P19-1586} {Low-resource deep
  entity resolution with transfer and active learning}.
\newblock In \emph{Proceedings of the 57th Annual Meeting of the Association
  for Computational Linguistics}, pages 5851--5861, Florence, Italy.
  Association for Computational Linguistics.

\bibitem[{Khalil et~al.(2019)Khalil, Kie{\l}czewski, Chouliaras, Keldibek, and
  Versteegh}]{projection/khalil-etal-2019-crossIntent}
Talaat Khalil, Kornel Kie{\l}czewski, Georgios~Christos Chouliaras, Amina
  Keldibek, and Maarten Versteegh. 2019.
\newblock \href {https://doi.org/10.18653/v1/D19-1676} {Cross-lingual intent
  classification in a low resource industrial setting}.
\newblock In \emph{Proceedings of the 2019 Conference on Empirical Methods in
  Natural Language Processing and the 9th International Joint Conference on
  Natural Language Processing (EMNLP-IJCNLP)}, pages 6419--6424, Hong Kong,
  China. Association for Computational Linguistics.

\bibitem[{Kiela et~al.(2018)Kiela, Wang, and Cho}]{emb/kiela-etal-2018-dynamic}
Douwe Kiela, Changhan Wang, and Kyunghyun Cho. 2018.
\newblock \href {https://doi.org/10.18653/v1/D18-1176} {Dynamic meta-embeddings
  for improved sentence representations}.
\newblock In \emph{Proceedings of the 2018 Conference on Empirical Methods in
  Natural Language Processing}, pages 1466--1477, Brussels, Belgium.
  Association for Computational Linguistics.

\bibitem[{Kim et~al.(2017)Kim, Kim, Sarikaya, and
  Fosler-Lussier}]{transfer/kim-etal-2017-cross-adversarial}
Joo-Kyung Kim, Young-Bum Kim, Ruhi Sarikaya, and Eric Fosler-Lussier. 2017.
\newblock \href {https://doi.org/10.18653/v1/D17-1302} {Cross-lingual transfer
  learning for {POS} tagging without cross-lingual resources}.
\newblock In \emph{Proceedings of the 2017 Conference on Empirical Methods in
  Natural Language Processing}, pages 2832--2838, Copenhagen, Denmark.
  Association for Computational Linguistics.

\bibitem[{Klie et~al.(2018)Klie, Bugert, Boullosa, de~Castilho, and
  Gurevych}]{distant/Klie18Inception}
Jan{-}Christoph Klie, Michael Bugert, Beto Boullosa, Richard~Eckart
  de~Castilho, and Iryna Gurevych. 2018.
\newblock \href {https://www.aclweb.org/anthology/C18-2002/} {The inception
  platform: Machine-assisted and knowledge-oriented interactive annotation}.
\newblock In \emph{{COLING} 2018, The 27th International Conference on
  Computational Linguistics: System Demonstrations, Santa Fe, New Mexico,
  August 20-26, 2018}, pages 5--9. Association for Computational Linguistics.

\bibitem[{Kobayashi(2018)}]{data-aug/kobayashi-2018-contextual}
Sosuke Kobayashi. 2018.
\newblock \href {https://doi.org/10.18653/v1/N18-2072} {Contextual
  augmentation: Data augmentation by words with paradigmatic relations}.
\newblock In \emph{Proceedings of the 2018 Conference of the North {A}merican
  Chapter of the Association for Computational Linguistics: Human Language
  Technologies, Volume 2 (Short Papers)}, pages 452--457, New Orleans,
  Louisiana. Association for Computational Linguistics.

\bibitem[{K{\"u}bler and Zhekova(2016)}]{app/kubler2016multilingual}
Sandra K{\"u}bler and Desislava Zhekova. 2016.
\newblock \href {https://onlinelibrary.wiley.com/doi/abs/10.1111/lnc3.12208}
  {Multilingual coreference resolution}.
\newblock \emph{Language and Linguistics Compass}, 10(11):614--631.

\bibitem[{Kumar et~al.(2020)Kumar, Choudhary, and
  Cho}]{data-aug/kumar2020transformer-aug-gen}
Varun Kumar, Ashutosh Choudhary, and Eunah Cho. 2020.
\newblock \href {https://www.aclweb.org/anthology/2020.lifelongnlp-1.3} {Data
  augmentation using pre-trained transformer models}.
\newblock In \emph{Proceedings of the 2nd Workshop on Life-long Learning for
  Spoken Language Systems}, pages 18--26, Suzhou, China. Association for
  Computational Linguistics.

\bibitem[{Lange et~al.(2019{\natexlab{a}})Lange, Adel, and
  Str{\"o}tgen}]{distant/lange2019pharma}
Lukas Lange, Heike Adel, and Jannik Str{\"o}tgen. 2019{\natexlab{a}}.
\newblock \href {https://doi.org/10.18653/v1/D19-5705} {{NLNDE}: Enhancing
  neural sequence taggers with attention and noisy channel for robust
  pharmacological entity detection}.
\newblock In \emph{Proceedings of The 5th Workshop on BioNLP Open Shared
  Tasks}, pages 26--32, Hong Kong, China. Association for Computational
  Linguistics.

\bibitem[{Lange et~al.(2019{\natexlab{b}})Lange, Adel, and
  Str{\"o}tgen}]{domain/lange2019nlnde}
Lukas Lange, Heike Adel, and Jannik Str{\"o}tgen. 2019{\natexlab{b}}.
\newblock \href {http://ceur-ws.org/Vol-2421/MEDDOCAN_paper_5.pdf} {Nlnde: The
  neither-language-nor-domain-experts' way of spanish medical document
  de-identification.}
\newblock In \emph{IberLEF@ SEPLN}, pages 671--678.

\bibitem[{Lange et~al.(2020{\natexlab{a}})Lange, Adel, and
  Str{\"o}tgen}]{emb/lange-etal-2020-choice}
Lukas Lange, Heike Adel, and Jannik Str{\"o}tgen. 2020{\natexlab{a}}.
\newblock \href {https://doi.org/10.18653/v1/2020.repl4nlp-1.13} {On the choice
  of auxiliary languages for improved sequence tagging}.
\newblock In \emph{Proceedings of the 5th Workshop on Representation Learning
  for NLP}, pages 95--102, Online. Association for Computational Linguistics.

\bibitem[{Lange et~al.(2020{\natexlab{b}})Lange, Adel, Str{\"o}tgen, and
  Klakow}]{emb/lange2020adversarial}
Lukas Lange, Heike Adel, Jannik Str{\"o}tgen, and Dietrich Klakow.
  2020{\natexlab{b}}.
\newblock \href {https://arxiv.org/pdf/2010.12305.pdf} {Adversarial learning of
  feature-based meta-embeddings}.
\newblock \emph{arXiv preprint arXiv:2010.12305}.

\bibitem[{Lange et~al.(2019{\natexlab{c}})Lange, Hedderich, and
  Klakow}]{distant/lange2019feature}
Lukas Lange, Michael~A. Hedderich, and Dietrich Klakow. 2019{\natexlab{c}}.
\newblock \href {https://doi.org/10.18653/v1/D19-1362} {Feature-dependent
  confusion matrices for low-resource {NER} labeling with noisy labels}.
\newblock In \emph{Proceedings of the 2019 Conference on Empirical Methods in
  Natural Language Processing and the 9th International Joint Conference on
  Natural Language Processing (EMNLP-IJCNLP)}, pages 3554--3559, Hong Kong,
  China. Association for Computational Linguistics.

\bibitem[{Lange et~al.(2020{\natexlab{c}})Lange, Iurshina, Adel, and
  Str{\"o}tgen}]{cross-ling/lange2020adversarial}
Lukas Lange, Anastasiia Iurshina, Heike Adel, and Jannik Str{\"o}tgen.
  2020{\natexlab{c}}.
\newblock \href {https://www.aclweb.org/anthology/2020.repl4nlp-1.14}
  {Adversarial alignment of multilingual models for extracting temporal
  expressions from text}.
\newblock In \emph{Proceedings of the 5th Workshop on Representation Learning
  for NLP}, pages 103--109, Online. Association for Computational Linguistics.

\bibitem[{Lauscher et~al.(2020)Lauscher, Ravishankar, Vuli{\'c}, and
  Glava{\v{s}}}]{lauscher2020zero}
Anne Lauscher, Vinit Ravishankar, Ivan Vuli{\'c}, and Goran Glava{\v{s}}. 2020.
\newblock \href {https://doi.org/10.18653/v1/2020.emnlp-main.363} {From zero to
  hero: {O}n the limitations of zero-shot language transfer with multilingual
  {T}ransformers}.
\newblock \emph{Proceedings of the 2020 Conference on Empirical Methods in
  Natural Language Processing (EMNLP)}, pages 4483--4499.

\bibitem[{Le and Titov(2019)}]{distant/le2019entityLinking}
Phong Le and Ivan Titov. 2019.
\newblock \href {https://doi.org/10.18653/v1/P19-1400} {Distant learning for
  entity linking with automatic noise detection}.
\newblock In \emph{Proceedings of the 57th Annual Meeting of the Association
  for Computational Linguistics}, pages 4081--4090, Florence, Italy.
  Association for Computational Linguistics.

\bibitem[{Lee and Hsiang(2020)}]{domain/lee2020patent}
Jieh-Sheng Lee and Jieh Hsiang. 2020.
\newblock \href
  {https://www.sciencedirect.com/science/article/pii/S0172219019300742} {Patent
  classification by fine-tuning bert language model}.
\newblock \emph{World Patent Information}, 61:101965.

\bibitem[{Lee et~al.(2020)Lee, Yoon, Kim, Kim, Kim, So, and
  Kang}]{domain/lee2020biobert}
Jinhyuk Lee, Wonjin Yoon, Sungdong Kim, Donghyeon Kim, Sunkyu Kim, Chan~Ho So,
  and Jaewoo Kang. 2020.
\newblock \href {https://doi.org/10.1093/bioinformatics/btz682} {Biobert: a
  pre-trained biomedical language representation model for biomedical text
  mining}.
\newblock \emph{Bioinformatics (Oxford, England)}, 36(4):1234—1240.

\bibitem[{Lee et~al.(2018)Lee, He, Zhang, and Yang}]{distant/Lee18FoodNoisy}
Kuang{-}Huei Lee, Xiaodong He, Lei Zhang, and Linjun Yang. 2018.
\newblock \href {https://doi.org/10.1109/CVPR.2018.00571} {Cleannet: Transfer
  learning for scalable image classifier training with label noise}.
\newblock In \emph{2018 {IEEE} Conference on Computer Vision and Pattern
  Recognition, {CVPR} 2018, Salt Lake City, UT, USA, June 18-22, 2018}, pages
  5447--5456. {IEEE} Computer Society.

\bibitem[{Li et~al.(2020)Li, Socher, and Hoi}]{distant/li20divide}
Junnan Li, Richard Socher, and Steven C.~H. Hoi. 2020.
\newblock \href {https://openreview.net/forum?id=HJgExaVtwr} {Dividemix:
  Learning with noisy labels as semi-supervised learning}.
\newblock In \emph{8th International Conference on Learning Representations,
  {ICLR} 2020, Addis Ababa, Ethiopia, April 26-30, 2020}. OpenReview.net.

\bibitem[{Li et~al.(2012)Li, Gra{\c{c}}a, and
  Taskar}]{distant/li-etal-2012-wiki}
Shen Li, Jo{\~a}o Gra{\c{c}}a, and Ben Taskar. 2012.
\newblock \href {https://www.aclweb.org/anthology/D12-1127} {{W}iki-ly
  supervised part-of-speech tagging}.
\newblock In \emph{Proceedings of the 2012 Joint Conference on Empirical
  Methods in Natural Language Processing and Computational Natural Language
  Learning}, pages 1389--1398, Jeju Island, Korea. Association for
  Computational Linguistics.

\bibitem[{Li et~al.(2017)Li, Wang, Li, Agustsson, and
  Gool}]{distant/Li17webvision}
Wen Li, Limin Wang, Wei Li, Eirikur Agustsson, and Luc~Van Gool. 2017.
\newblock \href {http://arxiv.org/abs/1708.02862} {Webvision database: Visual
  learning and understanding from web data}.
\newblock \emph{CoRR}, abs/1708.02862.

\bibitem[{Lin et~al.(2019)Lin, Chen, Lee, Li, Zhang, Xia, Rijhwani, He, Zhang,
  Ma, Anastasopoulos, Littell, and Neubig}]{cross-ling/lin-etal-2019-choosing}
Yu-Hsiang Lin, Chian-Yu Chen, Jean Lee, Zirui Li, Yuyan Zhang, Mengzhou Xia,
  Shruti Rijhwani, Junxian He, Zhisong Zhang, Xuezhe Ma, Antonios
  Anastasopoulos, Patrick Littell, and Graham Neubig. 2019.
\newblock \href {https://doi.org/10.18653/v1/P19-1301} {Choosing transfer
  languages for cross-lingual learning}.
\newblock In \emph{Proceedings of the 57th Annual Meeting of the Association
  for Computational Linguistics}, pages 3125--3135, Florence, Italy.
  Association for Computational Linguistics.

\bibitem[{Lison et~al.(2020)Lison, Barnes, Hubin, and
  Touileb}]{distant/lison-etal-2020-weak-supervision}
Pierre Lison, Jeremy Barnes, Aliaksandr Hubin, and Samia Touileb. 2020.
\newblock \href {https://www.aclweb.org/anthology/2020.acl-main.139} {Named
  entity recognition without labelled data: A weak supervision approach}.
\newblock In \emph{Proceedings of the 58th Annual Meeting of the Association
  for Computational Linguistics}, pages 1518--1533, Online. Association for
  Computational Linguistics.

\bibitem[{{Liu} et~al.(2019){Liu}, {Ma}, {Yang}, and
  {Yang}}]{survey/liu2019survey}
D.~{Liu}, N.~{Ma}, F.~{Yang}, and X.~{Yang}. 2019.
\newblock \href {https://ieeexplore.ieee.org/abstract/document/8969405} {A
  survey of low resource neural machine translation}.
\newblock In \emph{2019 4th International Conference on Mechanical, Control and
  Computer Engineering (ICMCCE)}, pages 39--393.

\bibitem[{Liu et~al.(2017)Liu, Qiu, and Huang}]{transfer/liu2017adversarial}
Pengfei Liu, Xipeng Qiu, and Xuanjing Huang. 2017.
\newblock \href {https://doi.org/10.18653/v1/P17-1001} {Adversarial multi-task
  learning for text classification}.
\newblock In \emph{Proceedings of the 55th Annual Meeting of the Association
  for Computational Linguistics, {ACL} 2017, Vancouver, Canada, July 30 -
  August 4, Volume 1: Long Papers}, pages 1--10. Association for Computational
  Linguistics.

\bibitem[{Liu et~al.(2019{\natexlab{a}})Liu, McCarthy, Vuli{\'c}, and
  Korhonen}]{cross-ling/liu2019investigating}
Qianchu Liu, Diana McCarthy, Ivan Vuli{\'c}, and Anna Korhonen.
  2019{\natexlab{a}}.
\newblock \href {https://doi.org/10.18653/v1/K19-1004} {Investigating
  cross-lingual alignment methods for contextualized embeddings with
  token-level evaluation}.
\newblock In \emph{Proceedings of the 23rd Conference on Computational Natural
  Language Learning (CoNLL)}.

\bibitem[{Liu et~al.(2019{\natexlab{b}})Liu, Ott, Goyal, Du, Joshi, Chen, Levy,
  Lewis, Zettlemoyer, and Stoyanov}]{pre-train/liu2019roberta}
Yinhan Liu, Myle Ott, Naman Goyal, Jingfei Du, Mandar Joshi, Danqi Chen, Omer
  Levy, Mike Lewis, Luke Zettlemoyer, and Veselin Stoyanov. 2019{\natexlab{b}}.
\newblock \href {https://arxiv.org/abs/1907.11692} {Roberta: A robustly
  optimized bert pretraining approach}.
\newblock \emph{arXiv preprint arXiv:1907.11692}.

\bibitem[{Longpre et~al.(2020)Longpre, Wang, and
  DuBois}]{data-aug/longpre2020dataaug-vs-pretrain}
Shayne Longpre, Yu~Wang, and Chris DuBois. 2020.
\newblock \href {https://doi.org/10.18653/v1/2020.findings-emnlp.394} {How
  effective is task-agnostic data augmentation for pretrained transformers?}
\newblock In \emph{Findings of the Association for Computational Linguistics:
  EMNLP 2020}, pages 4401--4411, Online. Association for Computational
  Linguistics.

\bibitem[{Loubser and Puttkammer(2020)}]{survey/loubser2020viability}
Melinda Loubser and Martin~J Puttkammer. 2020.
\newblock \href {https://www.mdpi.com/2078-2489/11/1/41} {Viability of neural
  networks for core technologies for resource-scarce languages}.
\newblock \emph{Information}, 11(1):41.

\bibitem[{Lozano et~al.(2013)Lozano, Farfan, and
  Cruz}]{app/lozano2013syntactic}
Jose Lozano, Waldir Farfan, and Juan Cruz. 2013.
\newblock \href
  {https://www.researchgate.net/profile/Jose-Lozano-31/publication/259265383_Syntactic_Analyzer_for_Quechua_Language/links/0deec52a9eb8d61f62000000/Syntactic-Analyzer-for-Quechua-Language.pdf}
  {Syntactic analyzer for quechua language}.

\bibitem[{Luo et~al.(2017)Luo, Feng, Wang, Zhu, Huang, Yan, and
  Zhao}]{distant/luo2017dynamicNoiseMatrix}
Bingfeng Luo, Yansong Feng, Zheng Wang, Zhanxing Zhu, Songfang Huang, Rui Yan,
  and Dongyan Zhao. 2017.
\newblock \href {https://doi.org/10.18653/v1/P17-1040} {Learning with noise:
  Enhance distantly supervised relation extraction with dynamic transition
  matrix}.
\newblock In \emph{Proceedings of the 55th Annual Meeting of the Association
  for Computational Linguistics (Volume 1: Long Papers)}, pages 430--439,
  Vancouver, Canada. Association for Computational Linguistics.

\bibitem[{Ma et~al.(2019)Ma, Yang, Liu, Li, Zhou, and
  Sun}]{data-aug/ma2019table2text}
Shuming Ma, Pengcheng Yang, Tianyu Liu, Peng Li, Jie Zhou, and Xu~Sun. 2019.
\newblock \href {https://doi.org/10.18653/v1/P19-1197} {Key fact as pivot: A
  two-stage model for low resource table-to-text generation}.
\newblock In \emph{Proceedings of the 57th Annual Meeting of the Association
  for Computational Linguistics}, pages 2047--2057, Florence, Italy.
  Association for Computational Linguistics.

\bibitem[{Mager et~al.(2018)Mager, Gutierrez-Vasques, Sierra, and
  Meza-Ruiz}]{survey/mager2018challenges}
Manuel Mager, Ximena Gutierrez-Vasques, Gerardo Sierra, and Ivan Meza-Ruiz.
  2018.
\newblock \href {https://www.aclweb.org/anthology/C18-1006} {Challenges of
  language technologies for the indigenous languages of the {A}mericas}.
\newblock In \emph{Proceedings of the 27th International Conference on
  Computational Linguistics}, pages 55--69, Santa Fe, New Mexico, USA.
  Association for Computational Linguistics.

\bibitem[{Magueresse et~al.(2020)Magueresse, Carles, and
  Heetderks}]{survey/magueresse2020low}
Alexandre Magueresse, Vincent Carles, and Evan Heetderks. 2020.
\newblock \href {https://arxiv.org/abs/2006.07264} {Low-resource languages: A
  review of past work and future challenges}.
\newblock \emph{arXiv preprint arXiv:2006.07264}.

\bibitem[{Mahajan et~al.(2018)Mahajan, Girshick, Ramanathan, He, Paluri, Li,
  Bharambe, and van~der Maaten}]{distant/mahajan18limits}
Dhruv Mahajan, Ross Girshick, Vignesh Ramanathan, Kaiming He, Manohar Paluri,
  Yixuan Li, Ashwin Bharambe, and Laurens van~der Maaten. 2018.
\newblock \href
  {https://openaccess.thecvf.com/content_ECCV_2018/html/Dhruv_Mahajan_Exploring_the_Limits_ECCV_2018_paper.html}
  {Exploring the limits of weakly supervised pretraining}.
\newblock \emph{Proceedings of the European Conference on Computer Vision
  (ECCV)}.

\bibitem[{Marasovi{\'c} et~al.(2016)Marasovi{\'c}, Zhou, Palmer, and
  Frank}]{projection/marasovic-etal-2016-modal}
Ana Marasovi{\'c}, Mengfei Zhou, Alexis Palmer, and Anette Frank. 2016.
\newblock \href {https://www.aclweb.org/anthology/2016.lilt-14.3} {Modal sense
  classification at large: Paraphrase-driven sense projection, semantically
  enriched classification models and cross-genre evaluations}.
\newblock In \emph{Linguistic Issues in Language Technology, Volume 14, 2016 -
  Modality: Logic, Semantics, Annotation, and Machine Learning}. CSLI
  Publications.

\bibitem[{Mart{\'\i}nez-Gil et~al.(2012)Mart{\'\i}nez-Gil,
  Zempoalteca-P{\'e}rez, Soancatl-Aguilar, de~Jes{\'u}s Estudillo-Ayala,
  Lara-Ram{\'\i}rez, and Alc{\'a}ntara-Santiago}]{app/martinez2012computer}
Carmen Mart{\'\i}nez-Gil, Alejandro Zempoalteca-P{\'e}rez, Venustiano
  Soancatl-Aguilar, Mar{\'\i}a de~Jes{\'u}s Estudillo-Ayala, Jos{\'e}~Edgar
  Lara-Ram{\'\i}rez, and Sayde Alc{\'a}ntara-Santiago. 2012.
\newblock \href
  {https://rcs.cic.ipn.mx/2012_47/Computer\%20Systems\%20for\%20Analysis\%20of\%20Nahuatl.pdf}
  {Computer systems for analysis of nahuatl.}
\newblock \emph{Res. Comput. Sci.}, 47:11--16.

\bibitem[{Mayer and Cysouw(2014)}]{projection/mayer-cysouw-2014-creating}
Thomas Mayer and Michael Cysouw. 2014.
\newblock \href
  {http://www.lrec-conf.org/proceedings/lrec2014/pdf/220_Paper.pdf} {Creating a
  massively parallel {B}ible corpus}.
\newblock In \emph{Proceedings of the Ninth International Conference on
  Language Resources and Evaluation ({LREC}'14)}, pages 3158--3163, Reykjavik,
  Iceland. European Language Resources Association (ELRA).

\bibitem[{Mayhew et~al.(2019)Mayhew, Chaturvedi, Tsai, and
  Roth}]{distant/mayhew2019partiallyNonSpeaker}
Stephen Mayhew, Snigdha Chaturvedi, Chen-Tse Tsai, and Dan Roth. 2019.
\newblock \href {https://doi.org/10.18653/v1/K19-1060} {Named entity
  recognition with partially annotated training data}.
\newblock In \emph{Proceedings of the 23rd Conference on Computational Natural
  Language Learning (CoNLL)}, pages 645--655, Hong Kong, China. Association for
  Computational Linguistics.

\bibitem[{Mayhew and Roth(2018)}]{distant/mayhew2018talen}
Stephen Mayhew and Dan Roth. 2018.
\newblock \href {https://doi.org/10.18653/v1/P18-4014} {{TALEN}: Tool for
  annotation of low-resource {EN}tities}.
\newblock In \emph{Proceedings of {ACL} 2018, System Demonstrations}, pages
  80--86, Melbourne, Australia. Association for Computational Linguistics.

\bibitem[{Mayhew et~al.(2017)Mayhew, Tsai, and
  Roth}]{projection/mayhew-etal-2017-cheap}
Stephen Mayhew, Chen-Tse Tsai, and Dan Roth. 2017.
\newblock \href {https://doi.org/10.18653/v1/D17-1269} {Cheap translation for
  cross-lingual named entity recognition}.
\newblock In \emph{Proceedings of the 2017 Conference on Empirical Methods in
  Natural Language Processing}, pages 2536--2545, Copenhagen, Denmark.
  Association for Computational Linguistics.

\bibitem[{Mekala et~al.(2020)Mekala, Zhang, and Shang}]{distant/mekala2020meta}
Dheeraj Mekala, Xinyang Zhang, and Jingbo Shang. 2020.
\newblock \href {https://doi.org/10.18653/v1/2020.emnlp-main.670} {{META}:
  Metadata-empowered weak supervision for text classification}.
\newblock In \emph{Proceedings of the 2020 Conference on Empirical Methods in
  Natural Language Processing (EMNLP)}, pages 8351--8361, Online. Association
  for Computational Linguistics.

\bibitem[{Melamud et~al.(2019)Melamud, Bornea, and
  Barker}]{pre-train/melamud2019}
Oren Melamud, Mihaela Bornea, and Ken Barker. 2019.
\newblock \href {https://doi.org/10.18653/v1/D19-1401} {Combining unsupervised
  pre-training and annotator rationales to improve low-shot text
  classification}.
\newblock In \emph{Proceedings of the 2019 Conference on Empirical Methods in
  Natural Language Processing and the 9th International Joint Conference on
  Natural Language Processing (EMNLP-IJCNLP)}, pages 3884--3893, Hong Kong,
  China. Association for Computational Linguistics.

\bibitem[{Mikolov et~al.(2013)Mikolov, Le, and
  Sutskever}]{multi-ling/mikolov2013exploiting}
Tomas Mikolov, Quoc~V. Le, and Ilya Sutskever. 2013.
\newblock \href {http://arxiv.org/abs/1309.4168} {Exploiting similarities among
  languages for machine translation}.
\newblock \emph{arXiv preprint arXiv:1309.4168}.

\bibitem[{Min et~al.(2020)Min, McCoy, Das, Pitler, and
  Linzen}]{data-aug/min-etal-2020-syntactic}
Junghyun Min, R.~Thomas McCoy, Dipanjan Das, Emily Pitler, and Tal Linzen.
  2020.
\newblock \href {https://doi.org/10.18653/v1/2020.acl-main.212} {Syntactic data
  augmentation increases robustness to inference heuristics}.
\newblock In \emph{Proceedings of the 58th Annual Meeting of the Association
  for Computational Linguistics}, pages 2339--2352, Online. Association for
  Computational Linguistics.

\bibitem[{Mintz et~al.(2009)Mintz, Bills, Snow, and
  Jurafsky}]{distant/mintz2009distant}
Mike Mintz, Steven Bills, Rion Snow, and Daniel Jurafsky. 2009.
\newblock \href {https://www.aclweb.org/anthology/P09-1113} {Distant
  supervision for relation extraction without labeled data}.
\newblock In \emph{Proceedings of the Joint Conference of the 47th Annual
  Meeting of the {ACL} and the 4th International Joint Conference on Natural
  Language Processing of the {AFNLP}}, pages 1003--1011, Suntec, Singapore.
  Association for Computational Linguistics.

\bibitem[{Morris et~al.(2020)Morris, Lifland, Yoo, Grigsby, Jin, and
  Qi}]{data-aug/morris2020textattack}
John Morris, Eli Lifland, Jin~Yong Yoo, Jake Grigsby, Di~Jin, and Yanjun Qi.
  2020.
\newblock \href {https://doi.org/10.18653/v1/2020.emnlp-demos.16}
  {{T}ext{A}ttack: A framework for adversarial attacks, data augmentation, and
  adversarial training in {NLP}}.
\newblock In \emph{Proceedings of the 2020 Conference on Empirical Methods in
  Natural Language Processing: System Demonstrations}, pages 119--126, Online.
  Association for Computational Linguistics.

\bibitem[{M{\"{u}}ller et~al.(2020)M{\"{u}}ller, Sagot, and
  Seddah}]{mueller20multitransfer}
Benjamin M{\"{u}}ller, Beno{\^{\i}}t Sagot, and Djam{\'{e}} Seddah. 2020.
\newblock \href {http://arxiv.org/abs/2005.00318} {Can multilingual language
  models transfer to an unseen dialect? {A} case study on north african
  arabizi}.
\newblock \emph{CoRR}, abs/2005.00318.

\bibitem[{M{\"u}{\"u}risep and Mutso(2005)}]{app/muurisep2005estsum}
Kaili M{\"u}{\"u}risep and Pilleriin Mutso. 2005.
\newblock \href
  {https://www.researchgate.net/profile/Kaili-Mueuerisep/publication/250152518_ESTSUM_-_ESTONIAN_NEWSPAPER_TEXTS_SUMMARIZER/links/5630aa3808ae1bdcebcf20be/ESTSUM-ESTONIAN-NEWSPAPER-TEXTS-SUMMARIZER.pdf}
  {Estsum-estonian newspaper texts summarizer}.
\newblock In \emph{Proceedings of The Second Baltic Conference on Human
  Language Technologies}, pages 311--316.

\bibitem[{Nekoto et~al.(2020)Nekoto, Marivate, Matsila, Fasubaa, Fagbohungbe,
  Akinola, Muhammad, Kabongo~Kabenamualu, Osei, Sackey, Niyongabo, Macharm,
  Ogayo, Ahia, Berhe, Adeyemi, Mokgesi-Selinga, Okegbemi, Martinus, Tajudeen,
  Degila, Ogueji, Siminyu, Kreutzer, Webster, Ali, Abbott, Orife, Ezeani,
  Dangana, Kamper, Elsahar, Duru, Kioko, Espoir, van Biljon, Whitenack,
  Onyefuluchi, Emezue, Dossou, Sibanda, Bassey, Olabiyi, Ramkilowan, {\"O}ktem,
  Akinfaderin, and Bashir}]{nekoto2020participatory}
Wilhelmina Nekoto, Vukosi Marivate, Tshinondiwa Matsila, Timi Fasubaa, Taiwo
  Fagbohungbe, Solomon~Oluwole Akinola, Shamsuddeen Muhammad, Salomon
  Kabongo~Kabenamualu, Salomey Osei, Freshia Sackey, Rubungo~Andre Niyongabo,
  Ricky Macharm, Perez Ogayo, Orevaoghene Ahia, Musie~Meressa Berhe, Mofetoluwa
  Adeyemi, Masabata Mokgesi-Selinga, Lawrence Okegbemi, Laura Martinus,
  Kolawole Tajudeen, Kevin Degila, Kelechi Ogueji, Kathleen Siminyu, Julia
  Kreutzer, Jason Webster, Jamiil~Toure Ali, Jade Abbott, Iroro Orife, Ignatius
  Ezeani, Idris~Abdulkadir Dangana, Herman Kamper, Hady Elsahar, Goodness Duru,
  Ghollah Kioko, Murhabazi Espoir, Elan van Biljon, Daniel Whitenack,
  Christopher Onyefuluchi, Chris~Chinenye Emezue, Bonaventure F.~P. Dossou,
  Blessing Sibanda, Blessing Bassey, Ayodele Olabiyi, Arshath Ramkilowan, Alp
  {\"O}ktem, Adewale Akinfaderin, and Abdallah Bashir. 2020.
\newblock \href {https://doi.org/10.18653/v1/2020.findings-emnlp.195}
  {Participatory research for low-resourced machine translation: A case study
  in {A}frican languages}.
\newblock In \emph{Findings of the Association for Computational Linguistics:
  EMNLP 2020}, pages 2144--2160, Online. Association for Computational
  Linguistics.

\bibitem[{Nivre et~al.(2020)Nivre, de~Marneffe, Ginter, Hajic, Manning,
  Pyysalo, Schuster, Tyers, and Zeman}]{defining/Nivre2020UD}
Joakim Nivre, Marie{-}Catherine de~Marneffe, Filip Ginter, Jan Hajic,
  Christopher~D. Manning, Sampo Pyysalo, Sebastian Schuster, Francis~M. Tyers,
  and Daniel Zeman. 2020.
\newblock \href {https://www.aclweb.org/anthology/2020.lrec-1.497/} {Universal
  dependencies v2: An evergrowing multilingual treebank collection}.
\newblock In \emph{Proceedings of The 12th Language Resources and Evaluation
  Conference, {LREC} 2020, Marseille, France, May 11-16, 2020}, pages
  4034--4043. European Language Resources Association.

\bibitem[{Nooralahzadeh et~al.(2019)Nooralahzadeh, L{\o}nning, and
  {\O}vrelid}]{distant/nooralahzadeh2019reinforcementDenoising}
Farhad Nooralahzadeh, Jan~Tore L{\o}nning, and Lilja {\O}vrelid. 2019.
\newblock \href {https://doi.org/10.18653/v1/D19-6125} {Reinforcement-based
  denoising of distantly supervised {NER} with partial annotation}.
\newblock In \emph{Proceedings of the 2nd Workshop on Deep Learning Approaches
  for Low-Resource NLP (DeepLo 2019)}, pages 225--233, Hong Kong, China.
  Association for Computational Linguistics.

\bibitem[{Norman et~al.(2019)Norman, Leeflang, Spijker, Kanoulas, and
  N{\'e}v{\'e}ol}]{distant/norman2019distantMedical}
Christopher Norman, Mariska Leeflang, Ren{\'e} Spijker, Evangelos Kanoulas, and
  Aur{\'e}lie N{\'e}v{\'e}ol. 2019.
\newblock \href {https://doi.org/10.18653/v1/W19-5012} {A distantly supervised
  dataset for automated data extraction from diagnostic studies}.
\newblock In \emph{Proceedings of the 18th BioNLP Workshop and Shared Task},
  pages 105--114, Florence, Italy. Association for Computational Linguistics.

\bibitem[{Olsson(2009)}]{active/olsson2009literature}
Fredrik Olsson. 2009.
\newblock \href {http://eprints.sics.se/3600/} {A literature survey of active
  machine learning in the context of natural language processing}.

\bibitem[{Onoe and Durrett(2019)}]{distant/onoe2019filteringRelabeling}
Yasumasa Onoe and Greg Durrett. 2019.
\newblock \href {https://doi.org/10.18653/v1/N19-1250} {Learning to denoise
  distantly-labeled data for entity typing}.
\newblock In \emph{Proceedings of the 2019 Conference of the North {A}merican
  Chapter of the Association for Computational Linguistics: Human Language
  Technologies, Volume 1 (Long and Short Papers)}, pages 2407--2417,
  Minneapolis, Minnesota. Association for Computational Linguistics.

\bibitem[{Opitz(2019)}]{distant/opitz19plausibility}
Juri Opitz. 2019.
\newblock \href
  {https://corpora.linguistik.uni-erlangen.de/data/konvens/proceedings/papers/KONVENS2019\_paper\_51.pdf}
  {Argumentative relation classification as plausibility ranking}.
\newblock In \emph{Proceedings of the 15th Conference on Natural Language
  Processing, {KONVENS} 2019, Erlangen, Germany, October 9-11, 2019}.

\bibitem[{Pajupuu et~al.(2016)Pajupuu, Altrov, and
  Pajupuu}]{app/pajupuu2016identifying}
Hille Pajupuu, Rene Altrov, and Jaan Pajupuu. 2016.
\newblock \href
  {https://www.researchgate.net/profile/Hille-Pajupuu/publication/303837298_Identifying_Polarity_in_Different_Text_Types/links/575711e308ae05c1ec16ce05/Identifying-Polarity-in-Different-Text-Types.pdf}
  {Identifying polarity in different text types}.
\newblock \emph{Folklore: Electronic Journal of Folklore}, 64:126--138.

\bibitem[{Pan and Yang(2009)}]{pan2009survey}
Sinno~Jialin Pan and Qiang Yang. 2009.
\newblock \href {https://ieeexplore.ieee.org/abstract/document/5288526/} {A
  survey on transfer learning}.
\newblock \emph{IEEE Transactions on knowledge and data engineering},
  22(10):1345--1359.

\bibitem[{Pan et~al.(2017)Pan, Zhang, May, Nothman, Knight, and
  Ji}]{eval/pan2017}
Xiaoman Pan, Boliang Zhang, Jonathan May, Joel Nothman, Kevin Knight, and Heng
  Ji. 2017.
\newblock \href {https://doi.org/10.18653/v1/P17-1178} {Cross-lingual name
  tagging and linking for 282 languages}.
\newblock In \emph{Proceedings of the 55th Annual Meeting of the Association
  for Computational Linguistics (Volume 1: Long Papers)}, pages 1946--1958,
  Vancouver, Canada. Association for Computational Linguistics.

\bibitem[{Parida and Motlicek(2019)}]{data-aug/parida2019abstract}
Shantipriya Parida and Petr Motlicek. 2019.
\newblock \href {https://doi.org/10.18653/v1/D19-1616} {Abstract text
  summarization: A low resource challenge}.
\newblock In \emph{Proceedings of the 2019 Conference on Empirical Methods in
  Natural Language Processing and the 9th International Joint Conference on
  Natural Language Processing (EMNLP-IJCNLP)}, pages 5994--5998, Hong Kong,
  China. Association for Computational Linguistics.

\bibitem[{Paul et~al.(2019)Paul, Singh, Hedderich, and
  Klakow}]{distant/paul-etal-2019-handling}
Debjit Paul, Mittul Singh, Michael~A. Hedderich, and Dietrich Klakow. 2019.
\newblock \href {https://doi.org/10.18653/v1/N19-3005} {Handling noisy labels
  for robustly learning from self-training data for low-resource sequence
  labeling}.
\newblock In \emph{Proceedings of the 2019 Conference of the North {A}merican
  Chapter of the Association for Computational Linguistics: Student Research
  Workshop}, pages 29--34, Minneapolis, Minnesota. Association for
  Computational Linguistics.

\bibitem[{Peng et~al.(2019)Peng, Xing, Zhang, Fu, and
  Huang}]{distant/peng-etal-2019-positive-unlabeled}
Minlong Peng, Xiaoyu Xing, Qi~Zhang, Jinlan Fu, and Xuanjing Huang. 2019.
\newblock \href {https://doi.org/10.18653/v1/P19-1231} {Distantly supervised
  named entity recognition using positive-unlabeled learning}.
\newblock In \emph{Proceedings of the 57th Annual Meeting of the Association
  for Computational Linguistics}, pages 2409--2419, Florence, Italy.
  Association for Computational Linguistics.

\bibitem[{Plank and Agi{\'c}(2018)}]{projection/plank-agic-2018-distant}
Barbara Plank and {\v{Z}}eljko Agi{\'c}. 2018.
\newblock \href {https://doi.org/10.18653/v1/D18-1061} {Distant supervision
  from disparate sources for low-resource part-of-speech tagging}.
\newblock In \emph{Proceedings of the 2018 Conference on Empirical Methods in
  Natural Language Processing}, pages 614--620, Brussels, Belgium. Association
  for Computational Linguistics.

\bibitem[{Plank et~al.(2016)Plank, S{\o}gaard, and
  Goldberg}]{defining/plank2016MultilingualPOS}
Barbara Plank, Anders S{\o}gaard, and Yoav Goldberg. 2016.
\newblock \href {https://doi.org/10.18653/v1/p16-2067} {Multilingual
  part-of-speech tagging with bidirectional long short-term memory models and
  auxiliary loss}.
\newblock In \emph{Proceedings of the 54th Annual Meeting of the Association
  for Computational Linguistics, {ACL} 2016, August 7-12, 2016, Berlin,
  Germany, Volume 2: Short Papers}. The Association for Computer Linguistics.

\bibitem[{Qian et~al.(2020)Qian, Chozhiyath~Raman, Li, and
  Popa}]{distant/qian-etal-2020-learning}
Kun Qian, Poornima Chozhiyath~Raman, Yunyao Li, and Lucian Popa. 2020.
\newblock \href {https://doi.org/10.18653/v1/2020.emnlp-main.517} {Learning
  structured representations of entity names using {A}ctive{L}earning and weak
  supervision}.
\newblock In \emph{Proceedings of the 2020 Conference on Empirical Methods in
  Natural Language Processing (EMNLP)}, pages 6376--6383, Online. Association
  for Computational Linguistics.

\bibitem[{Qiu et~al.(2020)Qiu, Sun, Xu, Shao, Dai, and
  Huang}]{survey/qiu2020pre}
Xipeng Qiu, Tianxiang Sun, Yige Xu, Yunfan Shao, Ning Dai, and Xuanjing Huang.
  2020.
\newblock \href {https://link.springer.com/article/10.1007/s11431-020-1647-3}
  {Pre-trained models for natural language processing: A survey}.
\newblock \emph{Science China Technological Sciences}, pages 1--26.

\bibitem[{Raffel et~al.(2020)Raffel, Shazeer, Roberts, Lee, Narang, Matena,
  Zhou, Li, and Liu}]{pre-train/raffel2019exploring}
Colin Raffel, Noam Shazeer, Adam Roberts, Katherine Lee, Sharan Narang, Michael
  Matena, Yanqi Zhou, Wei Li, and Peter~J Liu. 2020.
\newblock \href {https://www.jmlr.org/papers/volume21/20-074/20-074.pdf}
  {Exploring the limits of transfer learning with a unified text-to-text
  transformer}.
\newblock \emph{Journal of Machine Learning Research}, 21:1--67.

\bibitem[{Rahimi et~al.(2019)Rahimi, Li, and
  Cohn}]{transfer/rahimi-etal-2019-massively}
Afshin Rahimi, Yuan Li, and Trevor Cohn. 2019.
\newblock \href {https://doi.org/10.18653/v1/P19-1015} {Massively multilingual
  transfer for {NER}}.
\newblock In \emph{Proceedings of the 57th Annual Meeting of the Association
  for Computational Linguistics}, pages 151--164, Florence, Italy. Association
  for Computational Linguistics.

\bibitem[{Raiman and Miller(2017)}]{data-aug/raiman-miller-2017-globally}
Jonathan Raiman and John Miller. 2017.
\newblock \href {https://doi.org/10.18653/v1/D17-1111} {Globally normalized
  reader}.
\newblock In \emph{Proceedings of the 2017 Conference on Empirical Methods in
  Natural Language Processing}, pages 1059--1069, Copenhagen, Denmark.
  Association for Computational Linguistics.

\bibitem[{Ramponi and Plank(2020)}]{survey/ramponi2020neural}
Alan Ramponi and Barbara Plank. 2020.
\newblock \href {https://doi.org/10.18653/v1/2020.coling-main.603} {Neural
  unsupervised domain adaptation in {NLP}{---}{A} survey}.
\newblock \emph{Proceedings of the 28th International Conference on
  Computational Linguistics}, pages 6838--6855.

\bibitem[{Ratner et~al.(2017)Ratner, Bach, Ehrenberg, Fries, Wu, and
  R\'{e}}]{distant/ratner2019snorkel}
Alexander Ratner, Stephen~H. Bach, Henry Ehrenberg, Jason Fries, Sen Wu, and
  Christopher R\'{e}. 2017.
\newblock \href {https://doi.org/10.14778/3157794.3157797} {Snorkel: Rapid
  training data creation with weak supervision}.
\newblock \emph{Proc. VLDB Endow.}, 11(3):269–282.

\bibitem[{Rehbein and
  Ruppenhofer(2017)}]{distant/rehbein-ruppenhofer-2017-detecting}
Ines Rehbein and Josef Ruppenhofer. 2017.
\newblock \href {https://doi.org/10.18653/v1/P17-1107} {Detecting annotation
  noise in automatically labelled data}.
\newblock In \emph{Proceedings of the 55th Annual Meeting of the Association
  for Computational Linguistics (Volume 1: Long Papers)}, pages 1160--1170,
  Vancouver, Canada. Association for Computational Linguistics.

\bibitem[{Ren et~al.(2020)Ren, Li, Su, Kartchner, Mitchell, and
  Zhang}]{ren2020denoising}
Wendi Ren, Yinghao Li, Hanting Su, David Kartchner, Cassie Mitchell, and Chao
  Zhang. 2020.
\newblock \href {https://doi.org/10.18653/v1/2020.findings-emnlp.334}
  {Denoising multi-source weak supervision for neural text classification}.
\newblock In \emph{Findings of the Association for Computational Linguistics:
  EMNLP 2020}, pages 3739--3754, Online. Association for Computational
  Linguistics.

\bibitem[{Riedel et~al.(2010)Riedel, Yao, and
  McCallum}]{distant/riedel10multi-instance}
Sebastian Riedel, Limin Yao, and Andrew McCallum. 2010.
\newblock \href
  {https://link.springer.com/chapter/10.1007/978-3-642-15939-8_10} {Modeling
  relations and their mentions without labeled text}.
\newblock In \emph{Machine Learning and Knowledge Discovery in Databases},
  pages 148--163, Berlin, Heidelberg. Springer Berlin Heidelberg.

\bibitem[{Rogers et~al.(2021)Rogers, Kovaleva, and
  Rumshisky}]{survey/rogers2020primer}
Anna Rogers, Olga Kovaleva, and Anna Rumshisky. 2021.
\newblock \href
  {https://direct.mit.edu/tacl/article/doi/10.1162/tacl_a_00349/96482/A-Primer-in-BERTology-What-We-Know-About-How-BERT}
  {A primer in bertology: What we know about how bert works}.
\newblock \emph{Transactions of the Association for Computational Linguistics},
  8:842--866.

\bibitem[{Roth et~al.(2013)Roth, Barth, Wiegand, and
  Klakow}]{survey/roth2013survey}
Benjamin Roth, Tassilo Barth, Michael Wiegand, and Dietrich Klakow. 2013.
\newblock \href {https://dl.acm.org/doi/pdf/10.1145/2509558.2509571} {A survey
  of noise reduction methods for distant supervision}.
\newblock In \emph{Proceedings of the 2013 workshop on Automated knowledge base
  construction}, pages 73--78.

\bibitem[{Ruder(2019)}]{intro/ruder2019problems}
Sebastian Ruder. 2019.
\newblock \href {https://ruder.io/4-biggest-open-problems-in-nlp} {The 4
  biggest open problems in nlp}.

\bibitem[{Ruder et~al.(2019)Ruder, Vuli{\'c}, and
  S{\o}gaard}]{cross-ling/ruder2019survey}
Sebastian Ruder, Ivan Vuli{\'c}, and Anders S{\o}gaard. 2019.
\newblock \href {https://www.jair.org/index.php/jair/article/view/11640} {A
  survey of cross-lingual word embedding models}.
\newblock \emph{Journal of Artificial Intelligence Research}, 65:569--631.

\bibitem[{{\c{S}}ahin and Steedman(2018)}]{data-aug/sahin2018morphing}
G{\"o}zde~G{\"u}l {\c{S}}ahin and Mark Steedman. 2018.
\newblock \href {https://doi.org/10.18653/v1/D18-1545} {Data augmentation via
  dependency tree morphing for low-resource languages}.
\newblock In \emph{Proceedings of the 2018 Conference on Empirical Methods in
  Natural Language Processing}, pages 5004--5009, Brussels, Belgium.
  Association for Computational Linguistics.

\bibitem[{Schick et~al.(2020)Schick, Schmid, and
  Sch{\"u}tze}]{distant/schick2020automatically}
Timo Schick, Helmut Schmid, and Hinrich Sch{\"u}tze. 2020.
\newblock \href {https://doi.org/10.18653/v1/2020.coling-main.488}
  {Automatically identifying words that can serve as labels for few-shot text
  classification}.
\newblock In \emph{Proceedings of the 28th International Conference on
  Computational Linguistics}, pages 5569--5578, Barcelona, Spain (Online).
  International Committee on Computational Linguistics.

\bibitem[{Schick and Sch{\"{u}}tze(2020)}]{distant/schick2020exploiting}
Timo Schick and Hinrich Sch{\"{u}}tze. 2020.
\newblock \href {http://arxiv.org/abs/2001.07676} {Exploiting cloze questions
  for few-shot text classification and natural language inference}.
\newblock \emph{CoRR}, abs/2001.07676.

\bibitem[{Schick and Sch{\"u}tze(2020)}]{bert/schick2020s}
Timo Schick and Hinrich Sch{\"u}tze. 2020.
\newblock It's not just size that matters: Small language models are also
  few-shot learners.
\newblock \emph{arXiv preprint arXiv:2009.07118}.

\bibitem[{Schuster et~al.(2019)Schuster, Ram, Barzilay, and
  Globerson}]{cross-ling/schuster2019cross}
Tal Schuster, Ori Ram, Regina Barzilay, and Amir Globerson. 2019.
\newblock \href {https://doi.org/10.18653/v1/N19-1162} {Cross-lingual alignment
  of contextual word embeddings, with applications to zero-shot dependency
  parsing}.
\newblock In \emph{Proceedings of the 2019 Conference of the North {A}merican
  Chapter of the Association for Computational Linguistics: Human Language
  Technologies, Volume 1 (Long and Short Papers)}.

\bibitem[{Settles(2009)}]{active/settles2009survey}
Burr Settles. 2009.
\newblock \href {https://minds.wisconsin.edu/handle/1793/60660} {Active
  learning literature survey}.
\newblock Technical report, University of Wisconsin-Madison Department of
  Computer Sciences.

\bibitem[{Shi et~al.(2019)Shi, Xiao, and Niu}]{survey/shi2019brief}
Yong Shi, Yang Xiao, and Lingfeng Niu. 2019.
\newblock \href
  {https://link.springer.com/chapter/10.1007/978-3-030-22744-9_23} {A brief
  survey of relation extraction based on distant supervision}.
\newblock In \emph{International Conference on Computational Science}, pages
  293--303. Springer.

\bibitem[{Smirnova and Cudr{\'e}-Mauroux(2018)}]{survey/smirnova2018relation}
Alisa Smirnova and Philippe Cudr{\'e}-Mauroux. 2018.
\newblock Relation extraction using distant supervision: A survey.
\newblock \emph{ACM Computing Surveys (CSUR)}, 51(5):1--35.

\bibitem[{Steinberger(2012)}]{survey/steinberger2012survey}
Ralf Steinberger. 2012.
\newblock \href {https://doi.org/https://doi.org/10.1007/s10579-011-9165-9} {A
  survey of methods to ease the development of highly multilingual text mining
  applications}.
\newblock \emph{Language resources and evaluation}, 46(2):155--176.

\bibitem[{Strassel and Tracey(2016)}]{strassel-tracey-2016-lorelei}
Stephanie Strassel and Jennifer Tracey. 2016.
\newblock \href {https://www.aclweb.org/anthology/L16-1521} {{LORELEI} language
  packs: Data, tools, and resources for technology development in low resource
  languages}.
\newblock In \emph{Proceedings of the Tenth International Conference on
  Language Resources and Evaluation ({LREC}'16)}, pages 3273--3280,
  Portoro{\v{z}}, Slovenia. European Language Resources Association (ELRA).

\bibitem[{Surdeanu et~al.(2012)Surdeanu, Tibshirani, Nallapati, and
  Manning}]{distant/surdeanu-etal-2012-multi}
Mihai Surdeanu, Julie Tibshirani, Ramesh Nallapati, and Christopher~D. Manning.
  2012.
\newblock \href {https://www.aclweb.org/anthology/D12-1042} {Multi-instance
  multi-label learning for relation extraction}.
\newblock In \emph{Proceedings of the 2012 Joint Conference on Empirical
  Methods in Natural Language Processing and Computational Natural Language
  Learning}, pages 455--465, Jeju Island, Korea. Association for Computational
  Linguistics.

\bibitem[{T{\"a}ckstr{\"o}m et~al.(2013)T{\"a}ckstr{\"o}m, Das, Petrov,
  McDonald, and Nivre}]{projections/tackstrom-etal-2013-token}
Oscar T{\"a}ckstr{\"o}m, Dipanjan Das, Slav Petrov, Ryan McDonald, and Joakim
  Nivre. 2013.
\newblock \href {https://doi.org/10.1162/tacl_a_00205} {Token and type
  constraints for cross-lingual part-of-speech tagging}.
\newblock \emph{Transactions of the Association for Computational Linguistics},
  1:1--12.

\bibitem[{Tan et~al.(2018)Tan, Sun, Kong, Zhang, Yang, and Liu}]{tan2018survey}
Chuanqi Tan, Fuchun Sun, Tao Kong, Wenchang Zhang, Chao Yang, and Chunfang Liu.
  2018.
\newblock \href
  {https://link.springer.com/chapter/10.1007/978-3-030-01424-7_27} {A survey on
  deep transfer learning}.
\newblock In \emph{International conference on artificial neural networks},
  pages 270--279. Springer.

\bibitem[{Tiedemann(2012)}]{projection/tiedemann-2012-parallel}
J{\"o}rg Tiedemann. 2012.
\newblock \href
  {http://www.lrec-conf.org/proceedings/lrec2012/pdf/463_Paper.pdf} {Parallel
  data, tools and interfaces in {OPUS}}.
\newblock In \emph{Proceedings of the Eighth International Conference on
  Language Resources and Evaluation ({LREC}'12)}, pages 2214--2218, Istanbul,
  Turkey. European Language Resources Association (ELRA).

\bibitem[{Tkachenko et~al.(2013)Tkachenko, Petmanson, and
  Laur}]{app/tkachenko-etal-2013-named}
Alexander Tkachenko, Timo Petmanson, and Sven Laur. 2013.
\newblock \href {https://www.aclweb.org/anthology/W13-2412} {Named entity
  recognition in {E}stonian}.
\newblock In \emph{Proceedings of the 4th Biennial International Workshop on
  {B}alto-{S}lavic Natural Language Processing}, pages 78--83, Sofia, Bulgaria.
  Association for Computational Linguistics.

\bibitem[{Tracey and Strassel(2020)}]{app/tracey-strassel-2020-basic}
Jennifer Tracey and Stephanie Strassel. 2020.
\newblock \href {https://www.aclweb.org/anthology/2020.sltu-1.39} {Basic
  language resources for 31 languages (plus {E}nglish): The {LORELEI}
  representative and incident language packs}.
\newblock In \emph{Proceedings of the 1st Joint Workshop on Spoken Language
  Technologies for Under-resourced languages (SLTU) and Collaboration and
  Computing for Under-Resourced Languages (CCURL)}, pages 277--284, Marseille,
  France. European Language Resources association.

\bibitem[{Tsygankova et~al.(2020)Tsygankova, Marini, Mayhew, and
  Roth}]{distant/tsygankova2020nonspeaker}
Tatiana Tsygankova, Francesca Marini, Stephen Mayhew, and Dan Roth. 2020.
\newblock \href {http://arxiv.org/abs/2006.09627} {Building low-resource ner
  models using non-speaker annotation}.
\newblock \emph{CoRR}, abs/2006.09627.

\bibitem[{Tukur et~al.(2019)Tukur, Umar, and Muhammad}]{app/tukur2019tagging}
Aminu Tukur, Kabir Umar, and Anas~Saidu Muhammad. 2019.
\newblock \href {https://ieeexplore.ieee.org/abstract/document/9043198/}
  {Tagging part of speech in hausa sentences}.
\newblock In \emph{2019 15th International Conference on Electronics, Computer
  and Computation (ICECCO)}, pages 1--6. IEEE.

\bibitem[{Vania et~al.(2019)Vania, Kementchedjhieva, S{\o}gaard, and
  Lopez}]{data-aug/vania2019parsing}
Clara Vania, Yova Kementchedjhieva, Anders S{\o}gaard, and Adam Lopez. 2019.
\newblock \href {https://doi.org/10.18653/v1/D19-1102} {A systematic comparison
  of methods for low-resource dependency parsing on genuinely low-resource
  languages}.
\newblock In \emph{Proceedings of the 2019 Conference on Empirical Methods in
  Natural Language Processing and the 9th International Joint Conference on
  Natural Language Processing (EMNLP-IJCNLP)}, pages 1105--1116, Hong Kong,
  China. Association for Computational Linguistics.

\bibitem[{Vaswani et~al.(2017)Vaswani, Shazeer, Parmar, Uszkoreit, Jones,
  Gomez, Kaiser, and Polosukhin}]{pre-train/vaswani2017attention}
Ashish Vaswani, Noam Shazeer, Niki Parmar, Jakob Uszkoreit, Llion Jones,
  Aidan~N Gomez, \L~ukasz Kaiser, and Illia Polosukhin. 2017.
\newblock \href
  {http://papers.nips.cc/paper/7181-attention-is-all-you-need.pdf} {Attention
  is all you need}.
\newblock In I.~Guyon, U.~V. Luxburg, S.~Bengio, H.~Wallach, R.~Fergus,
  S.~Vishwanathan, and R.~Garnett, editors, \emph{Advances in Neural
  Information Processing Systems 30}, pages 5998--6008. Curran Associates, Inc.

\bibitem[{Wang et~al.(2019)Wang, Liu, Li, Yang, and
  Li}]{distant/wang2019sentiment}
Hao Wang, Bing Liu, Chaozhuo Li, Yan Yang, and Tianrui Li. 2019.
\newblock \href {https://doi.org/10.18653/v1/D19-1655} {Learning with noisy
  labels for sentence-level sentiment classification}.
\newblock In \emph{Proceedings of the 2019 Conference on Empirical Methods in
  Natural Language Processing and the 9th International Joint Conference on
  Natural Language Processing (EMNLP-IJCNLP)}, pages 6286--6292, Hong Kong,
  China. Association for Computational Linguistics.

\bibitem[{Wei and Zou(2019)}]{data-aug/wei-zou-2019-eda}
Jason Wei and Kai Zou. 2019.
\newblock \href {https://doi.org/10.18653/v1/D19-1670} {{EDA}: Easy data
  augmentation techniques for boosting performance on text classification
  tasks}.
\newblock In \emph{Proceedings of the 2019 Conference on Empirical Methods in
  Natural Language Processing and the 9th International Joint Conference on
  Natural Language Processing (EMNLP-IJCNLP)}, pages 6382--6388, Hong Kong,
  China. Association for Computational Linguistics.

\bibitem[{Weiss et~al.(2016)Weiss, Khoshgoftaar, and Wang}]{weiss2016survey}
Karl Weiss, Taghi~M Khoshgoftaar, and DingDing Wang. 2016.
\newblock \href
  {https://journalofbigdata.springeropen.com/articles/10.1186/s40537-016-0043-6}
  {A survey of transfer learning}.
\newblock \emph{Journal of Big data}, 3(1):9.

\bibitem[{Wilson and Cook(2020)}]{survey/wilson2018domain}
Garrett Wilson and Diane~J Cook. 2020.
\newblock \href {https://dl.acm.org/doi/abs/10.1145/3400066} {A survey of
  unsupervised deep domain adaptation}.
\newblock \emph{ACM Transactions on Intelligent Systems and Technology (TIST)},
  11(5):1--46.

\bibitem[{Wisniewski et~al.(2014)Wisniewski, P{\'e}cheux, Gahbiche-Braham, and
  Yvon}]{projection/wisniewski-etal-2014-cross-lingual}
Guillaume Wisniewski, Nicolas P{\'e}cheux, Souhir Gahbiche-Braham, and
  Fran{\c{c}}ois Yvon. 2014.
\newblock \href {https://doi.org/10.3115/v1/D14-1187} {Cross-lingual
  part-of-speech tagging through ambiguous learning}.
\newblock In \emph{Proceedings of the 2014 Conference on Empirical Methods in
  Natural Language Processing ({EMNLP})}, pages 1779--1785, Doha, Qatar.
  Association for Computational Linguistics.

\bibitem[{Wu and Dredze(2020)}]{multi-ling/wu-dredze-2020-languages}
Shijie Wu and Mark Dredze. 2020.
\newblock \href {https://doi.org/10.18653/v1/2020.repl4nlp-1.16} {Are all
  languages created equal in multilingual {BERT}?}
\newblock In \emph{Proceedings of the 5th Workshop on Representation Learning
  for NLP}, pages 120--130, Online. Association for Computational Linguistics.

\bibitem[{Xiao et~al.(2015)Xiao, Xia, Yang, Huang, and
  Wang}]{distant/xiao15clothing1m}
Tong Xiao, Tian Xia, Yi~Yang, Chang Huang, and Xiaogang Wang. 2015.
\newblock \href {https://doi.org/10.1109/CVPR.2015.7298885} {Learning from
  massive noisy labeled data for image classification}.
\newblock In \emph{{IEEE} Conference on Computer Vision and Pattern
  Recognition, {CVPR} 2015, Boston, MA, USA, June 7-12, 2015}, pages
  2691--2699. {IEEE} Computer Society.

\bibitem[{Xie et~al.(2020)Xie, Dai, Hovy, Luong, and
  Le}]{data-aug/xie2020backtranslation}
Qizhe Xie, Zihang Dai, Eduard Hovy, Minh-Thang Luong, and Quoc~V. Le. 2020.
\newblock \href
  {https://proceedings.neurips.cc/paper/2020/hash/44feb0096faa8326192570788b38c1d1-Abstract.html}
  {Unsupervised data augmentation for consistency training}.
\newblock In \emph{Advances in Neural Information Processing Systems 33: Annual
  Conference on Neural Information Processing Systems 2020, NeurIPS 2020,
  December 6-12, 2020, virtual}.

\bibitem[{Xu et~al.(2020)Xu, Liu, Shu, and Yu}]{domain/xu2020dombert}
Hu~Xu, Bing Liu, Lei Shu, and Philip Yu. 2020.
\newblock \href {https://doi.org/10.18653/v1/2020.findings-emnlp.156}
  {{D}om{BERT}: Domain-oriented language model for aspect-based sentiment
  analysis}.
\newblock \emph{Findings of the Association for Computational Linguistics:
  EMNLP 2020}, pages 1725--1731.

\bibitem[{Yang et~al.(2018)Yang, Chen, Li, He, and
  Zhang}]{distant/yang2018distantPartialAnnotationReinforcement}
Yaosheng Yang, Wenliang Chen, Zhenghua Li, Zhengqiu He, and Min Zhang. 2018.
\newblock \href {https://www.aclweb.org/anthology/C18-1183} {Distantly
  supervised {NER} with partial annotation learning and reinforcement
  learning}.
\newblock In \emph{Proceedings of the 27th International Conference on
  Computational Linguistics}, pages 2159--2169, Santa Fe, New Mexico, USA.
  Association for Computational Linguistics.

\bibitem[{Yang et~al.(2019)Yang, Wu, Yang, Xu, and
  Li}]{defining/yang2019responseGeneration}
Ze~Yang, Wei Wu, Jian Yang, Can Xu, and Zhoujun Li. 2019.
\newblock \href {https://doi.org/10.18653/v1/D19-1197} {Low-resource response
  generation with template prior}.
\newblock In \emph{Proceedings of the 2019 Conference on Empirical Methods in
  Natural Language Processing and the 9th International Joint Conference on
  Natural Language Processing (EMNLP-IJCNLP)}, pages 1886--1897, Hong Kong,
  China. Association for Computational Linguistics.

\bibitem[{Yarowsky et~al.(2001)Yarowsky, Ngai, and
  Wicentowski}]{projection/yarowsky-etal-2001-inducing}
David Yarowsky, Grace Ngai, and Richard Wicentowski. 2001.
\newblock \href {https://www.aclweb.org/anthology/H01-1035} {Inducing
  multilingual text analysis tools via robust projection across aligned
  corpora}.
\newblock In \emph{Proceedings of the First International Conference on Human
  Language Technology Research}.

\bibitem[{Yasunaga et~al.(2018)Yasunaga, Kasai, and
  Radev}]{data-aug/yasunaga-etal-2018-robust}
Michihiro Yasunaga, Jungo Kasai, and Dragomir Radev. 2018.
\newblock \href {https://doi.org/10.18653/v1/N18-1089} {Robust multilingual
  part-of-speech tagging via adversarial training}.
\newblock In \emph{Proceedings of the 2018 Conference of the North {A}merican
  Chapter of the Association for Computational Linguistics: Human Language
  Technologies, Volume 1 (Long Papers)}, pages 976--986, New Orleans,
  Louisiana. Association for Computational Linguistics.

\bibitem[{Ye et~al.(2019)Ye, Liu, Zhang, and Ren}]{distant/ye2019shift}
Qinyuan Ye, Liyuan Liu, Maosen Zhang, and Xiang Ren. 2019.
\newblock \href {https://doi.org/10.18653/v1/D19-1397} {Looking beyond label
  noise: Shifted label distribution matters in distantly supervised relation
  extraction}.
\newblock In \emph{Proceedings of the 2019 Conference on Empirical Methods in
  Natural Language Processing and the 9th International Joint Conference on
  Natural Language Processing (EMNLP-IJCNLP)}, pages 3841--3850, Hong Kong,
  China. Association for Computational Linguistics.

\bibitem[{Younes et~al.(2020)Younes, Souissi, Achour, and
  Ferchichi}]{survey/younes2020language}
Jihene Younes, Emna Souissi, Hadhemi Achour, and Ahmed Ferchichi. 2020.
\newblock \href
  {https://link.springer.com/article/10.1007\%2Fs10579-020-09490-9} {Language
  resources for maghrebi arabic dialects’ nlp: a survey}.
\newblock \emph{LANGUAGE RESOURCES AND EVALUATION}.

\bibitem[{Yu et~al.(2018)Yu, Guo, Yi, Chang, Potdar, Cheng, Tesauro, Wang, and
  Zhou}]{transfer/yu2018metrics}
Mo~Yu, Xiaoxiao Guo, Jinfeng Yi, Shiyu Chang, Saloni Potdar, Yu~Cheng, Gerald
  Tesauro, Haoyu Wang, and Bowen Zhou. 2018.
\newblock \href {https://doi.org/10.18653/v1/N18-1109} {Diverse few-shot text
  classification with multiple metrics}.
\newblock In \emph{Proceedings of the 2018 Conference of the North {A}merican
  Chapter of the Association for Computational Linguistics: Human Language
  Technologies, Volume 1 (Long Papers)}, pages 1206--1215, New Orleans,
  Louisiana. Association for Computational Linguistics.

\bibitem[{Yude(2011)}]{survey/yude2011brief}
BI~Yude. 2011.
\newblock \href {http://en.cnki.com.cn/Article_en/CJFDTotal-MESS201106022.htm}
  {A brief survey of korean natural language processing research}.
\newblock \emph{Journal of Chinese Information Processing}, 6.

\bibitem[{Zhang et~al.(2019{\natexlab{a}})Zhang, Zhang, and
  Fu}]{projection/zhang-etal-2019-cross}
Meishan Zhang, Yue Zhang, and Guohong Fu. 2019{\natexlab{a}}.
\newblock \href {https://doi.org/10.18653/v1/D19-1092} {Cross-lingual
  dependency parsing using code-mixed {T}ree{B}ank}.
\newblock In \emph{Proceedings of the 2019 Conference on Empirical Methods in
  Natural Language Processing and the 9th International Joint Conference on
  Natural Language Processing (EMNLP-IJCNLP)}, pages 997--1006, Hong Kong,
  China. Association for Computational Linguistics.

\bibitem[{Zhang et~al.(2019{\natexlab{b}})Zhang, Westerfield, Shim, Bingham,
  Fabbri, Hu, Verma, and Radev}]{cross-ling/zhang2019improving}
Rui Zhang, Caitlin Westerfield, Sungrok Shim, Garrett Bingham, Alexander
  Fabbri, William Hu, Neha Verma, and Dragomir Radev. 2019{\natexlab{b}}.
\newblock \href {https://doi.org/10.18653/v1/P19-1306} {Improving low-resource
  cross-lingual document retrieval by reranking with deep bilingual
  representations}.
\newblock In \emph{Proceedings of the 57th Annual Meeting of the Association
  for Computational Linguistics}, pages 3173--3179, Florence, Italy.
  Association for Computational Linguistics.

\bibitem[{Zheng et~al.(2019)Zheng, Han, Lin, Yu, Chen, Huang, Liu, and
  Xu}]{distant/zheng2019rules}
Shun Zheng, Xu~Han, Yankai Lin, Peilin Yu, Lu~Chen, Ling Huang, Zhiyuan Liu,
  and Wei Xu. 2019.
\newblock \href {https://doi.org/10.18653/v1/P19-1137} {{DIAG}-{NRE}: A neural
  pattern diagnosis framework for distantly supervised neural relation
  extraction}.
\newblock In \emph{Proceedings of the 57th Annual Meeting of the Association
  for Computational Linguistics}, pages 1419--1429, Florence, Italy.
  Association for Computational Linguistics.

\bibitem[{Zhou et~al.(2019)Zhou, Zhang, Jin, Zhu, Fang, Goh, and
  Kwok}]{transfer/zhou2019dualAdversarial}
Joey~Tianyi Zhou, Hao Zhang, Di~Jin, Hongyuan Zhu, Meng Fang, Rick Siow~Mong
  Goh, and Kenneth Kwok. 2019.
\newblock \href {https://doi.org/10.18653/v1/P19-1336} {Dual adversarial neural
  transfer for low-resource named entity recognition}.
\newblock In \emph{Proceedings of the 57th Annual Meeting of the Association
  for Computational Linguistics}, pages 3461--3471, Florence, Italy.
  Association for Computational Linguistics.

\bibitem[{Zhou et~al.(2015)Zhou, Frank, Friedrich, and
  Palmer}]{projection/zhou-etal-2015-semantically}
Mengfei Zhou, Anette Frank, Annemarie Friedrich, and Alexis Palmer. 2015.
\newblock \href {https://doi.org/10.18653/v1/W15-2705} {Semantically enriched
  models for modal sense classification}.
\newblock In \emph{Proceedings of the First Workshop on Linking Computational
  Models of Lexical, Sentential and Discourse-level Semantics}, pages 44--53,
  Lisbon, Portugal. Association for Computational Linguistics.

\bibitem[{Zhu et~al.(2019)Zhu, Heinzerling, Vuli{\'c}, Strube, Reichart, and
  Korhonen}]{emb/zhu2019subword}
Yi~Zhu, Benjamin Heinzerling, Ivan Vuli{\'c}, Michael Strube, Roi Reichart, and
  Anna Korhonen. 2019.
\newblock \href {https://doi.org/10.18653/v1/K19-1021} {On the importance of
  subword information for morphological tasks in truly low-resource languages}.
\newblock In \emph{Proceedings of the 23rd Conference on Computational Natural
  Language Learning (CoNLL)}, pages 216--226, Hong Kong, China. Association for
  Computational Linguistics.

\end{thebibliography}
\bibliographystyle{acl_natbib}

\appendix

\begin{table*}
\footnotesize
\centering
\begin{tabular}{lll}
\toprule
& Low-resource surveys & \newcite{survey/cieri2016selection} , \newcite{survey/magueresse2020low} \\
\midrule
\multirow{9}{*}{\rotatebox{90}{\textit{Method-specific}}}
& Active learning & \newcite{active/olsson2009literature}, \newcite{active/settles2009survey}, \newcite{active/aggarwal2014survey} \\
& Distant supervision & \newcite{survey/roth2013survey}, \newcite{survey/smirnova2018relation}, \newcite{survey/shi2019brief}. \\
& Unsupervised domain adaptation & \newcite{survey/wilson2018domain}, \newcite{survey/ramponi2020neural} \\
& Meta-Learning & \newcite{survey/hospedales2020meta} \\
& Multilingual transfer & \newcite{survey/steinberger2012survey}, \newcite{cross-ling/ruder2019survey} \\
& LM pre-training & \newcite{survey/rogers2020primer}, \newcite{survey/qiu2020pre} \\
& Machine translation & \newcite{survey/liu2019survey} \\
& Label noise handling & \newcite{survey/frenay2013classification}, \newcite{survey/algan2019image} \\
& Transfer learning & \newcite{pan2009survey}, \newcite{weiss2016survey}, \newcite{tan2018survey} \\
\midrule
\multirow{5}{*}{\rotatebox{90}{\textit{Language-}}} 
& African languages & \newcite{survey/grover2010overview}, \newcite{survey/de2011introduction}  \\
& Arabic languages & \newcite{survey/al2018deep}, \newcite{survey/guellil2019arabic}, \newcite{survey/younes2020language} \\
& American languages & \newcite{survey/mager2018challenges} \\
& South-Asian languages & \newcite{survey/daud2017urdu}, \newcite{survey/banik2019bengali}, \newcite{survey/harish2020comprehensive} \\
& East-Asian languages & \newcite{survey/yude2011brief} \\
\bottomrule
\end{tabular}
\caption{Overview of existing surveys on low-resource topics.}
\label{tab:surveys}
\end{table*}

\section{Existing Surveys on Low-Resource Topics and Languages}
There is a growing body of task- and language-specific surveys concerning low-resource topics. We list these surveys in Table~\ref{tab:surveys} as a starting point for a more in-depth reading regarding specific topics.

\section{Complexity of Tasks}\label{app:tasks}
While a large number of labeled resources for English are available for many popular NLP tasks, this is not the case for the majority of low-resource languages. To measure (and visualize as done in Figure \ref{fig:applications} in the main paper) which applications are accessible to speakers of low-resource languages we examined resources for six different languages, ranging from high- to low-resource languages for a fixed set of tasks of varying complexity, ranging from basic tasks, such as tokenization, to higher-level tasks, such as question answering.
For this short study, we have chosen the following languages. The number of speakers are the combined L1 and L2 speakers according to \newcite{intro/Ethnologue2019}.
\begin{itemize}
\item[(1)] English: The most high-resource language according to the common view and literature in the NLP community.
\item[(2)] Yoruba: An African language, which is spoken by about 40 million speakers and contained in the EXTREME benchmark \cite{eval/hu2020xtreme}. 
Even with that many speakers, this language is often considered as a low-resource language and it is still discussed whether this language is also endangered \cite{app/fabuni2005yoruba}.
\item[(3)] Hausa: An African language with over 60 million speakers. It is not covered in EXTREME or the universal dependencies project \cite{defining/Nivre2020UD}.
\item[(4)] Quechua: A language family encompassing about 8 million speakers, mostly in Peru.
\item[(5)] Nahuatl and (6) Estonian: Both have between 1 and 2 million speakers, but are spoken in very different regions (North America \& Europe).
\end{itemize}

All speaker numbers according to \newcite{intro/Ethnologue2019} reflecting the total number of users (L1 + L2). The tasks were chosen from a list of popular NLP tasks\footnote{\url{https://en.wikipedia.org/wiki/Natural\_language\_processing\#Common\_NLP\_Tasks}}.
We selected two tasks for the lower-lever groups and three tasks for the higher-level groups, which reflects the application diversity with increasing complexity.
Table~\ref{tab:tasks-covered} shows which tasks were addressed for each language.

Word segmentation, lemmatization, part-of-speech tagging, sentence breaking and (semantic) parsing are covered for Yoruba and Estonian by treebanks from the universal dependencies project \cite{defining/Nivre2020UD}. 
Cusco Quechua is listed as an upcoming language in the UD project, but no treebank is accessible at this moment.
The WikiAnn corpus for named entity recognition \cite{eval/pan2017} has resources and tools for NER and sentence breaking for all six languages.
Lemmatization resources for Nahuatl were developed by \newcite{app/martinez2012computer} and \newcite{app/lozano2013syntactic} developed resources for part-of-speech tagging, tokenization and parsing of Quechuan. 
The CoNLL conference and SIGMORPHON organized two shared tasks for morphological reinflection which provided lemmatization resources for many languages, including Quechuan \cite{survey/conll2018universalMorphologicalReinflection}.

Basic resources for simple semantic role labeling and entity linking were developed during the LORELEI program for many low-resource languages \cite{strassel-tracey-2016-lorelei,app/tracey-strassel-2020-basic}, including resources for Yoruba and Hausa (even though the latter "fell short" according to the authors).
Estonian coreference resolution is targeted by \newcite{app/kubler2016multilingual}, but the available resources are very limited. Estonian sentiment is done by \newcite{app/pajupuu2016identifying}.
All languages are covered by the multilingual fasttext embeddings \cite{emb/bojanowski2017enriching} and byte-pair-encoding embeddings \cite{emb/heinzerling2018bpemb}. Yoruba, Hausa and Estonian are covered by mBERT or XLM-RoBERTa as well.

Text summarization is done for Estonian by \newcite{app/muurisep2005estsum} and for Hausa by \newcite{app/bashir2017automatic}.
The EXTREME benchmark \cite{eval/hu2020xtreme} covers question answering and natural language inference tasks for Yoruba and Estonian (besides NER, POS tagging and more).
Publicly available systems for optical character recognition support all six languages \cite{app/hakro2016multilingual}.
All these tasks are supported for the English language as well, and most often, the English datasets are many times larger and of much higher quality. Some of the previously mentioned datasets were automatically translated, as in the EXTREME benchmark for several languages. As outlined in the main paper, we do not claim that all tasks marked in the Table yield high-performance model, but we instead indicate if any resources or models can be found for a language.

\begin{landscape}

\setlength{\tabcolsep}{3pt}
\begin{table}
\footnotesize

\begin{tabular}{llccccc} \toprule
Group & Task
& Yoruba & Hausa  & Quechuan & Nahuatl & Estonian \\ \midrule
 & Num-Speakers
 & 40 mil. & 60 mil. & 8 mil. & 1.7 mil. & 1.3 mil.\\ \midrule
\multirow{2}{*}{Text processing} 
 & Word segmentation
 & \ding{51} & \ding{51} & \ding{51} & \ding{51} & \ding{51} \\
 & Optical character recognition
 & \citet{app/hakro2016multilingual} & \citet{app/hakro2016multilingual} & \citet{app/hakro2016multilingual} & \citet{app/hakro2016multilingual} & \citet{app/hakro2016multilingual} \\ \midrule
\multirow{2}{*}{Morphological analysis} 
& Lemmatization / Stemming
& \citet{survey/conll2018universalMorphologicalReinflection} & \citet{survey/conll2018universalMorphologicalReinflection} & \citet{survey/conll2018universalMorphologicalReinflection} & \citet{app/martinez2012computer} & \citet{survey/conll2018universalMorphologicalReinflection} \\
 & Part-of-Speech tagging
 & \citet{defining/Nivre2020UD} & \citet{app/tukur2019tagging} & \newcite{app/lozano2013syntactic} & \ding{55} & \citet{defining/Nivre2020UD} \\ \midrule
\multirow{2}{*}{Syntactic analysis} 
 & Sentence breaking
 & \ding{51} & \ding{51} & \ding{51} & \ding{51} & \ding{51} \\
 & Parsing
 & \citet{defining/Nivre2020UD} & \ding{55} & \citet{defining/Nivre2020UD} & \ding{55} & \citet{defining/Nivre2020UD} \\ \midrule
\multirow{2}{*}{Distributional semantics} 
 & Word embeddings
 & FT, BPEmb & FT, BPEmb & FT, BPEmb & FT, BPEmb & FT, BPEmb \\
 & Transformer models
 & mBERT & XLM-R & \ding{55} & \ding{55} & mBERT, XLM-R \\ \midrule
\multirow{2}{*}{Lexical semantics} 
 & Named entity recognition
 & \citet{survey/adelani2020distant} & \citet{survey/adelani2020distant} & \citet{eval/pan2017} & \citet{eval/pan2017} & \citet{app/tkachenko-etal-2013-named} \\
 & Sentiment analysis
 & \ding{55} & \ding{55} & \ding{55} & \ding{55} & \citet{app/pajupuu2016identifying} \\ \midrule
\multirow{3}{*}{Relational semantics} 
 & Relationship extraction
 & \ding{55} & \ding{55} & \ding{55} & \ding{55} & \ding{55} \\
 & Semantic Role Labelling
 & \citet{app/tracey-strassel-2020-basic} & \citet{app/tracey-strassel-2020-basic} & \ding{55} & \ding{55} & \ding{55} \\
 & Semantic Parsing
 & \citet{defining/Nivre2020UD} & \ding{55} & \ding{55} & \ding{55} & \citet{defining/Nivre2020UD} \\ \midrule
\multirow{3}{*}{Discourse} 
 & Coreference resolution
 & \ding{55} & \ding{55} & \ding{55} & \ding{55} & \citet{app/kubler2016multilingual} \\
 & Discourse analysis
 & \ding{55} & \ding{55} & \ding{55} & \ding{55} &  \\
 & Textual entailment
 & \citet{eval/hu2020xtreme} & \ding{55} & \ding{55} & \ding{55} & \citet{eval/hu2020xtreme} \\ \midrule
\multirow{3}{*}{Higher-level NLP} 
 & Text summarization
 & \ding{55} & \citet{app/bashir2017automatic} & \ding{55} & \ding{55} & \citet{app/muurisep2005estsum} \\
 & Dialogue management 
 & \ding{55} & \ding{55} & \ding{55} & \ding{55} &  \ding{55} \\
 & Question answering (QA)
 & \citet{eval/hu2020xtreme} & \ding{55} & \ding{55} & \ding{55} & \citet{eval/hu2020xtreme} \\ \midrule
 & SUM
 & 13 & 10 & 8 & 6 & 15\\ \bottomrule
\end{tabular}

\caption{Overview of tasks covered by six different languages. Note that this list is non-exhaustive and due to space reasons we only give one reference per language and task. }
\label{tab:tasks-covered}
\setlength{\tabcolsep}{6pt}
\end{table}
\end{landscape}

\end{document}